\documentclass[9pt,journal,cspaper,compsoc]{IEEEtran}

\ifCLASSOPTIONcompsoc
\else
  \usepackage{cite}
\fi

\ifCLASSINFOpdf
\else
  \usepackage[dvips]{graphicx}
\fi
\usepackage{epsfig}

\usepackage[cmex10]{amsmath}
\usepackage{array}

\usepackage{mdwmath}
\usepackage{mdwtab}
\usepackage{eqparbox}
\usepackage{multirow}

\usepackage{url}
\usepackage{amssymb}
\usepackage{amsthm,url,balance}
\usepackage{color}
\usepackage[T1]{fontenc}

\usepackage{algorithm}
\usepackage{algorithmic}
\usepackage[ruled,vlined,algo2e]{algorithm2e}
\providecommand{\SetAlgoLined}{\SetLine}

\usepackage{dsfont}	
\usepackage{wrapfig}
\usepackage{footnote}

\newcommand{\ie}{{\it i.e. }}
\newcommand{\etal}{{\it et al. }}
\newcommand{\eg}{{\it e.g. }}
\newcommand{\vs}{{\it vs. }}
\newcommand{\etc}{{\it etc.}}
\newcommand{\wrt}{{\it w.r.t. }}

\newtheorem*{lemma*}{Lemma}
\DeclareMathOperator*{\argmin}{arg\,min}
\DeclareMathOperator*{\argmax}{arg\,max}
\newcommand{\rmnum}[1]{\romannumeral #1}

\begin{document}
%
\title{Sequential Optimization for Efficient High-Quality Object Proposal Generation}

\author{Ziming Zhang, Yun Liu, Xi Chen, Yanjun Zhu, Ming-Ming Cheng, Venkatesh Saligrama, and Philip H.S. Torr
\IEEEcompsocitemizethanks{
	\IEEEcompsocthanksitem Dr. Z. Zhang is with Mitsubishi Electric Research Laboratories (MERL), Cambridge, MA 02139-1955, U.S. E-mail: zzhang@merl.com \protect	
	\IEEEcompsocthanksitem Y. Liu and Prof. M-M. Cheng are with CCCE \& CS, Nankai University, Tianjin 300071, China. E-mail: nk12csly@mail.nankai.edu.cn, cmm@nankai.edu.cn \protect
	\IEEEcompsocthanksitem X. Chen and Y. Zhu are with the School of Automation, Huazhong University of Science and Technology, Wuhan 430074, China. E-mail: \{chenxihust, yjzhu\}@hust.edu.cn, \protect
	\IEEEcompsocthanksitem Prof. V. Saligrama is with the Department of Electrical and Computer Engineering, Boston University, Boston, MA 02215, US. E-mail: srv@bu.edu \protect
	\IEEEcompsocthanksitem Prof. P.H.S. Torr is with the Department of Engineering Science, University of Oxford, Oxford OX1 3PJ, UK. E-mail: philip.torr@eng.ox.ac.uk \protect
}
\thanks{}}

%
%

\markboth{IEEE Transaction on Pattern Analysis and Machine Intelligence}%
{Zhang \etal: Sequential Optimization for Efficient High-Quality Object Proposal Generation}
%


\IEEEcompsoctitleabstractindextext{%
\begin{abstract}
We are motivated by the need for a generic object proposal generation algorithm which achieves good balance between object detection recall, proposal localization quality and computational efficiency. We propose a novel object proposal algorithm, {\em BING++}, which inherits the virtue of good computational efficiency of BING \cite{BingObj2014} but significantly improves its proposal localization quality. At high level we formulate the problem of object proposal generation from a novel probabilistic perspective, 
based on which our BING++ manages to improve the localization quality by employing edges and segments to estimate object boundaries and update the proposals sequentially. We 
propose learning the parameters efficiently by searching for approximate solutions in a quantized parameter space for complexity reduction. We demonstrate the generalization of BING++ with the same fixed parameters across different object classes and datasets.
Empirically our BING++ can run at {\em half} speed of BING on CPU, 
but significantly improve the localization quality 
by 18.5\% and 16.7\% on both VOC2007 and Microhsoft COCO datasets, respectively. Compared with other state-of-the-art approaches, BING++ can achieve comparable performance, but run significantly faster. 
\end{abstract}

}

\maketitle

\IEEEdisplaynotcompsoctitleabstractindextext

%
\IEEEpeerreviewmaketitle

\section{Introduction}
\IEEEPARstart{G}eneric object proposal generation arises as a critical standalone preprocessing step in many applications such as object recognition \cite{wei2014cnn} and detection \cite{girshick2014rich}, and consequently has attracted significant attention. Object proposal generation can be broadly measured using three metrics: (a) {\em Detection Recall} (DR) \cite{BingObj2014, zhang2011proposal, zhang2016object}, which is the ratio between the number of correctly detected objects and the total number of objects in the dataset; (b) {\em Proposal Localization Quality} in terms of average best overlap (ABO) for each object instance in each class, and corresponding mean average best overlap (MABO) across all the classes \cite{Uijlings13}; (c) {\em Computational Efficiency} (CE). In this paper, we are interested in developing new algorithms to provide a small set of windows ({\it i.e.} bounding boxes) in images with high DR, high localization quality (especially for MABO), and high CE.


\begin{figure}[t]
\begin{minipage}[b]{0.49\linewidth}
 \begin{center}
 \centerline{\includegraphics[width=\columnwidth]{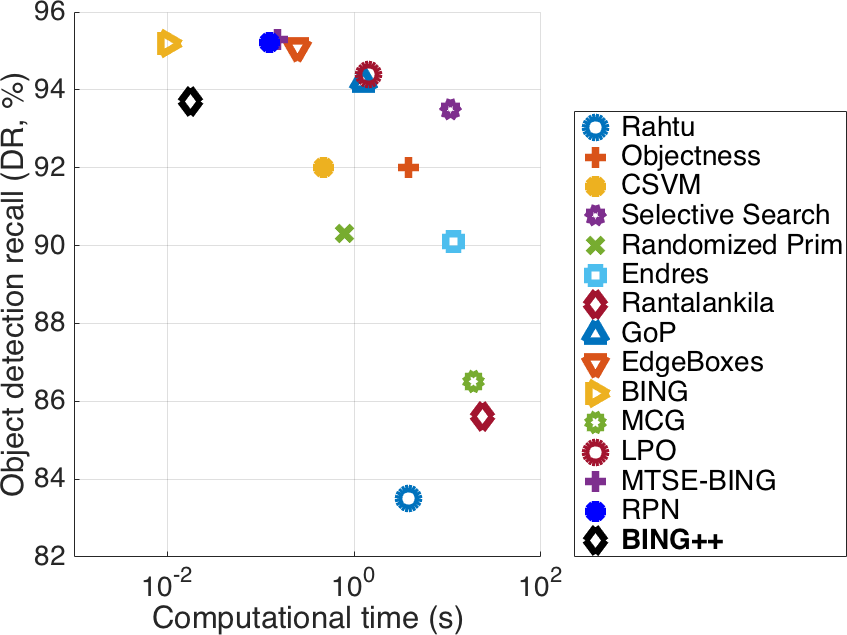}}
 \centerline{\footnotesize{(a) DR vs. computational time}}
 \end{center}
\end{minipage}
\begin{minipage}[b]{0.49\linewidth}
\begin{center}
\centerline{\includegraphics[width=\columnwidth]{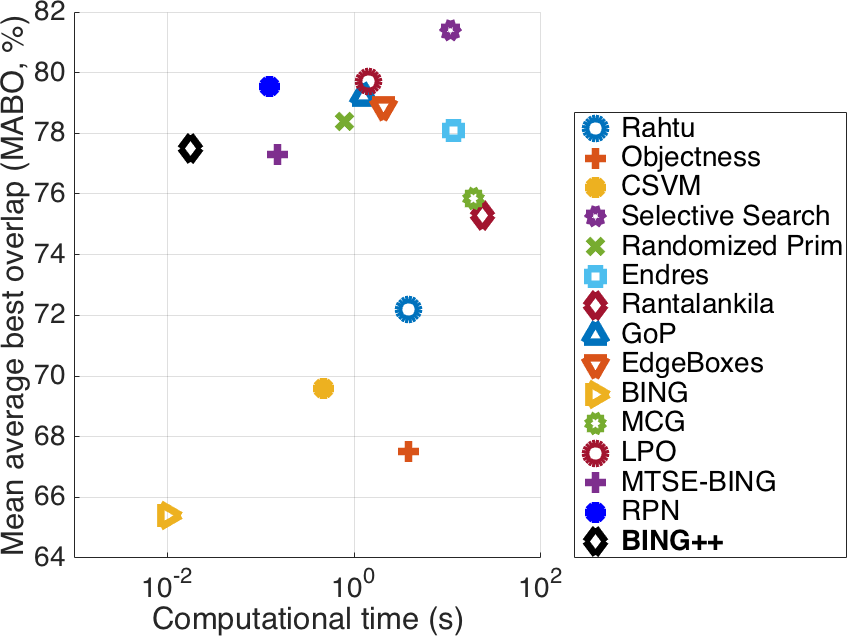}}
 \centerline{\footnotesize{(b) MABO vs. computational time}}
\end{center} 
\end{minipage}
\vspace{-3mm}
\caption{\footnotesize{Comparison of generic object proposal methods on VOC2007 test dataset \cite{pascal-voc-2007} with at most 1,000 proposals per image and intersection-over-union (IoU) threshold equal to 0.5.
All the competing results are produced by public code (see Table \ref{tab:2} and \ref{tab:ABO} in Section \ref{sec:exp} for more details).}}\label{fig:comparison}
\vspace{-3mm}
\end{figure}

In recent years while many object proposal generation algorithms have been proposed, existing methods do not appear to achieve good balance between DR, MABO and CE. Fig. \ref{fig:comparison} depicts inherent tradeoffs in DR, MABO and CE among different proposal algorithms. We can see clearly that, for instance, BING \cite{BingObj2014} is computationally efficient but has poor localization quality, while selective search \cite{Uijlings13} generates good proposals but is computationally inefficient. Our perspective here is that computational efficiency has to be an important consideration in developing algorithms since object proposal generation is typically a preprocessing step.
Based on this reasoning, we propose a novel object proposal algorithm, {\em BING++}, which is an extension of our previous work \cite{BingObj2014,zhang2011proposal,zhang2016object}.

\begin{wrapfigure}{t}{0.45\linewidth} 
	\vskip-2ex
	\centering
	\includegraphics[width = \linewidth]{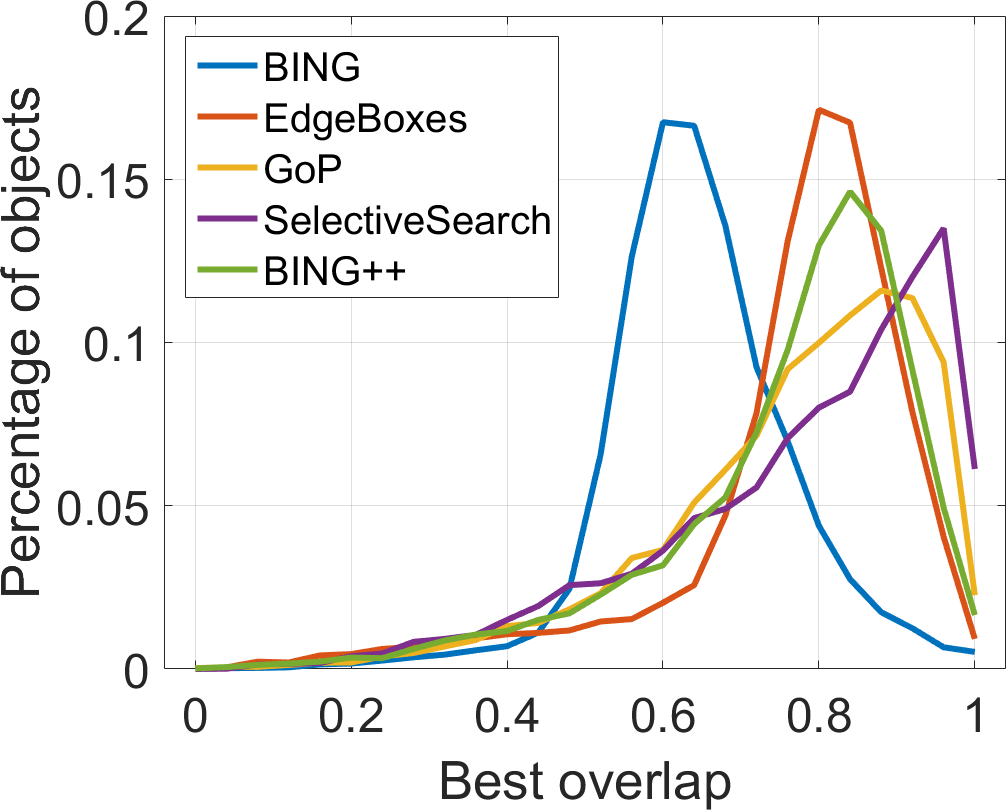}
	\vspace{-3mm}
	\caption{\footnotesize{Best overlap (BO) statistical comparison using at most 1,000 proposals per image and IoU threshold 0.5 on VOC2007 test dataset.
		}}
	\label{fig:abo-stats}
\end{wrapfigure}

BING \cite{BingObj2014} has been demonstrated as a very efficient object proposal algorithm. Its basic idea is to first train linear filters for each so-called {\em quantized scale/aspect-ratio} (or quantized window size) \cite{zhang2011proposal, zhang2016object} using simple binary gradient features, and then learn another global linear filter to rank bounding boxes from each quantized scale/aspect-ratio and output proposals from the top list. The quantization scheme guarantees to map every possible object scale/aspect-ratio to at least one of the {\em predefined and fixed} quantized scales/aspect-ratios. As stated in \cite{zhang2016object}, this quantization scheme reduces the proposal searching space logarithmically, leading to very high computational efficiency. However, this step also leads to significant degradation in proposal quality in practice.

\begin{figure*}[t]
\begin{minipage}[b]{0.245\linewidth}
 \begin{center}
 \centerline{\includegraphics[width=.8\linewidth]{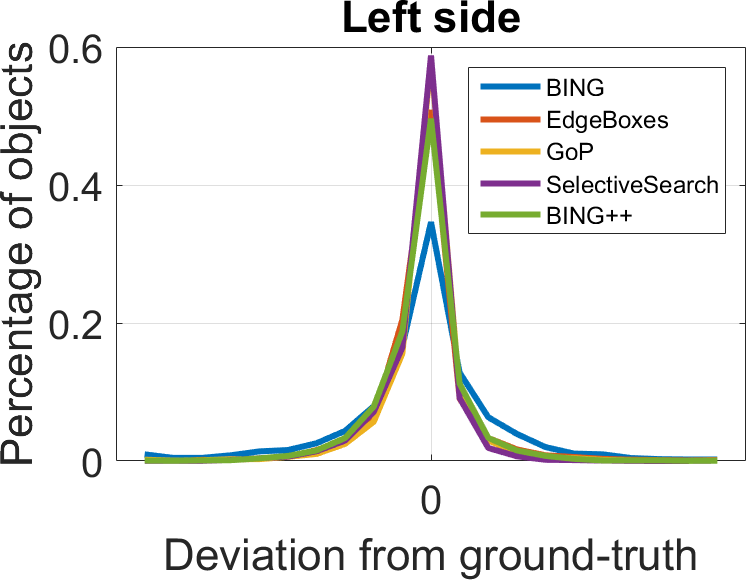}}
 \end{center}
\end{minipage}
\begin{minipage}[b]{0.245\linewidth}
\begin{center}
\centerline{\includegraphics[width=.8\linewidth]{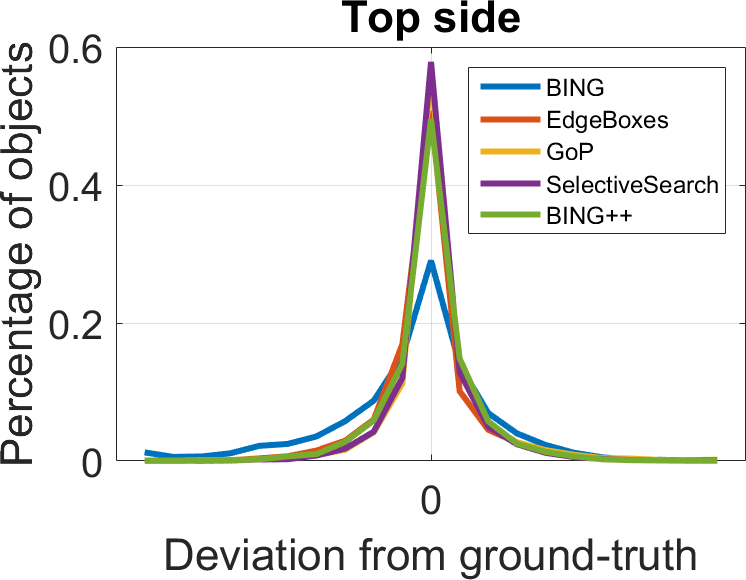}}
\end{center} 
\end{minipage}
\begin{minipage}[b]{0.245\linewidth}
 \begin{center}
 \centerline{\includegraphics[width=.8\linewidth]{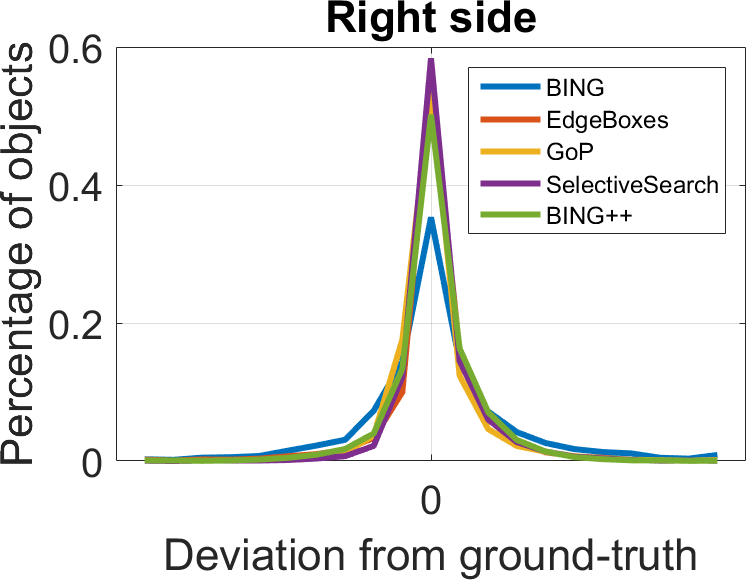}}
 \end{center}
\end{minipage}
\begin{minipage}[b]{0.245\linewidth}
\begin{center}
\centerline{\includegraphics[width=.8\linewidth]{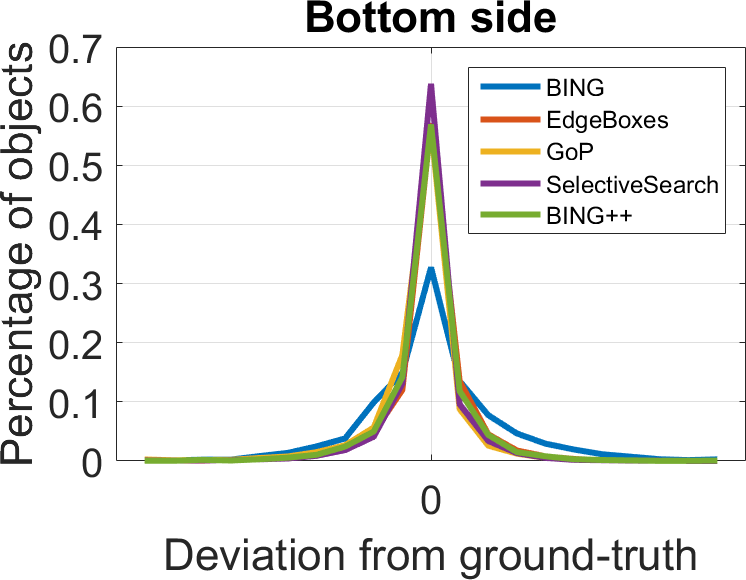}}
\end{center} 
\end{minipage}
\vspace{-9mm}
\caption{\footnotesize{Statistical comparison based on percentage of objects \vs best proposal deviation from the ground-truth bounding box per object with at most 1,000 proposals per image and IoU threshold 0.5 on VOC2007 test dataset.}}
\label{fig:stats}
\vspace{-3mm}
\end{figure*}

To see this, here we will show some statistics about the proposal quality on VOC2007 test set. The behavior on either training or test set is similar. We first point to the best overlap (BO) statistics in Fig.~\ref{fig:abo-stats}, where we notice that the localization quality of BING proposals (over the dataset) is mediocre, because there is a clear leftward drift in BING's distribution relative to other methods. This is indicative of poor proposal localization quality of BING. In order to see how proposals drift from the ground-truth, we point to comparison between the boundary deviation statistics based on percentage of objects and the best proposal deviation from the ground-truth bounding box per object in Fig.~\ref{fig:stats}. Ideally a Dirac delta distribution in this context is preferable, and the closer the distribution to Dirac delta, the better the proposal algorithm in terms of localization quality. Compared to other competitors, BING performs worse because its distributions appear to have heavier tails on both sides. This indicates that BING is agnostic between choosing larger or smaller proposals for the ground-truth.

On the other hand since BING has high DR, we conclude that {\em the quantization scheme in BING leads to poor proposal localization quality}. We infer this based on the view that BING does not allow the proposals to be adaptive to the object boundaries. In contrast, the methods that fully utilize either edge information (\eg \cite{zitnickedge}) or segments/superpixels (\eg \cite{Uijlings13}) perform better than BING, but run much slower. \\

\noindent
\underline{\em Contributions:} In this paper we propose a {\em fast yet accurate} object proposal algorithm, BING++\footnote{The code is available at \url{https://zimingzhang.wordpress.com/}.}, which inherits the virtues of BING, \ie computational efficiency and objectness scores, but significantly improves its proposal localization quality. For instance, on VOC2007 BING++ achieves 77.5\% in terms of MABO with significant improvement of {\bf 18.5\%} over BING within only about {\bf 3ms} on our server with two INTEL XEON E5 2696v2 CPU@2.50GHz. To our best knowledge {\em BING++ is the fastest object proposal algorithm among those that produce state-of-the-art quality}. 

We first propose a novel probabilistic perspective for understanding the problem of object proposal generation, where we reveal the nature of recursive updating mechanism of proposals by alternating optimization of estimating object boundaries and updating proposals accordingly. This probabilistic view can be served as the theoretical justification of our BING++ (as well as some other proposal algorithms such as \cite{wang2015improving}). We learn the corresponding parameters efficiently by searching for approximate solutions in a quantized parameter space for complexity reduction. We demonstrate the robustness of such learned parameters across different datasets. In reality we utilize edges and segments sequentially to estimate object boundaries as guide to update proposals further. In such way BING++ manages to improve the localization quality recursively in a coarse-to-fine manner. 
We test BING++ on VOC2007 and Microsoft COCO \cite{lin2014microsoft} datasets, and consistently achieve better trade-off between DR, MABO, and CE, compared with many other object proposal generation algorithms.


\subsection{BING for Objectness Measure}
\label{ssec:BING-overview}

Objects are typically considered as stand-alone things with well-defined closed boundaries and centers \cite{forsyth1996finding}. When resizing windows corresponding to real world objects to a small fixed size (\eg $8 \times 8$), the norm (\ie magnitude) of the corresponding image gradients becomes a good discriminative feature, due to the little variation that closed boundaries could present in such abstracted views. Also inspired by the ability of human vision system which efficiently perceives objects before identifying them \cite{55ARP/Teuber_physiological}, we introduce a simple 64-dim {\em normed gradient (NG)} feature as well as its approximation, \ie {\em binarized normed gradient (BING)} feature, for efficiently capturing the objectness of an image window.

To find generic objects in an image, we scan over predefined \emph{quantized window sizes} (scales and aspect ratios) \cite{zhang2011proposal, zhang2016object}. Each window is scored with a linear model $\mathbf{w}\in \mathbb{R}^{64}$:
\begin{equation}\label{equ:score}
s_l = \langle\mathbf{w},\mathbf{g}_l\rangle, \; l = (i,x,y),
\end{equation}
where $s_l$, $\mathbf{g}_l$, $l$, $i$ and $(x,y)$ are filter score, NG feature, location, size and position of a window respectively, and $\langle\cdot,\cdot\rangle$ denotes the inner product operator of two vectors. Using non-maximal suppression (NMS), we select a small set of proposals from each size $i$. Some sizes (\eg $10 \times 500$) are less likely than others to contain an object instance (\eg $100 \times 100$). Thus we define the objectness score (\ie calibrated filter score), $c_l$, as
\begin{equation}\label{equ:calibration}
  c_l = v_i s_l + t_i,
\end{equation}
where $v_i, t_i \in \mathbb{R}$ are separately learned coefficient and a bias terms for each quantized size $i$. Note that calibration based on Eq. \ref{equ:calibration}, although very fast, is only required when re-ranking the small set of final proposals.

To make use of recent advantages in model binarization approximation 
\cite{hare2012efficient,szhengFacialFG2013}, 
we propose an accelerated version of NG features, \ie BING, 
to speed up the feature extraction and testing process. 
Our learned linear model $\mathbf{w}\in\mathbb{R}^{64}$ can be approximated 
with a set of basis vectors 
$\textbf{w} \approx \sum_{j=1}^{N_w} \beta_j \mathbf{a}_j$ using 
\cite{hare2012efficient}, 
where $N_w$ denotes the number of basis vectors, 
$\mathbf{a}_j \in \{-1, 1\}^{64}$ denotes a basis vector, 
and $\beta_j\in\mathbb{R}$ denotes the corresponding coefficient. 
By further representing each $\mathbf{a}_j$ using a binary vector and its 
complement: $\mathbf{a}_j = \mathbf{a}_j^+ - \overline{\mathbf{a}_j^+}$, 
where $\mathbf{a}_j^+ \in \{0,1\}^{64}$, 
a binarized feature $\mathbf{b}$ could be tested using fast 
{\sc bitwise and} and {\sc bit count} operations 
(see \cite{hare2012efficient}), 
\begin{equation}\label{equ:modelApro}
  \langle\mathbf{w}, \mathbf{b} \rangle \approx \sum\nolimits^{N_w}_{j=1} \beta_j(2 \langle \mathbf{a}_j^+, \mathbf{b} \rangle  - |\mathbf{b}|).
\end{equation}

We approximate the NG values (each saved as a {\sc byte} value) using the top $N_g$ binary bits of the {\sc byte} values. Thus, a 64-dim NG feature $\mathbf{g}_l$ can be approximated by $N_g$ (BING) features as  
\begin{equation}\label{equ:BING}
  \mathbf{g}_l = \sum\nolimits^{N_g}_{k=1} 2^{8-k}\mathbf{b}_{k,l},
\end{equation}
where $\mathbf{b}_{k,l}, \forall k, \forall l$ is a binarized feature. Note that these BING features have different weights based on their bit positions in {\sc byte} values. Accordingly the filter score in Eq.~\ref{equ:score} of a window corresponding to BING feature $\mathbf{b}_{k,l}$ can be efficiently tested using:
\begin{equation}\label{equ:scoreFast}
  s_{l} \approx \sum\nolimits^{N_w}_{j=1} \beta_j \sum\nolimits^{N_g}_{k=1} C_{j,k},
\end{equation}
where $C_{j,k} = 2^{8-k} (2\langle\mathbf{a}_j^+, \mathbf{b}_{k,l}\rangle - |\mathbf{b}_{k,l}|)$ 
can be tested using fast {\sc bitwise} and {\sc popcnt sse} operators. For more details please refer to our paper \cite{BingObj2014}. 

\subsection{Related Work}

Object proposal generation algorithms for images in the literature can be further categorized into three groups, in general, as follows:

{\em (\rmnum{1}) Segmentation/Superpixel based algorithms:} In fact most proposal generation algorithms fall into this group. For instance, objectness measure \cite{Alexe2012pami} combines saliency, color, edges, and superpixels to score the windows, and then samples bounding boxes with high scores as object proposals. Based on \cite{Alexe2012pami}, Rahtu \etal \cite{Rahtu_iccv11} proposed another cascaded method, where the proposal candidates are sampled from super-pixels based on a prior object localization distribution and then ranked using structured learning with learned features. Further in \cite{Blaschko2013b}, Blaschko \etal investigated the effect of the NMS step in \cite{Rahtu_iccv11} to improve the performance. Uijlings \etal \cite{Uijlings13} proposed selective search by combining the strength of both an exhaustive search and segmentation and being guided by the image structure. Manen \etal \cite{Manen2013iccv} proposed a randomized Prim algorithm on the superpixel connectivity graphs. Endres and Hoiem \cite{Endres2012} proposed ranking the a set of segments using structured learning based on various cues. Kr{\"a}henb{\"u}hl and Koltun \cite{krahenbuhl2014geodesic} proposed identifying critical level sets in geodesic distance transforms as proposals, and in \cite{kk-lpo-15} they proposed learning ensembles of classifiers for generating proposals. There are several methods based on energy minimization, such as constrained parametric min-cut \cite{Carreira2012, Humayun2015}, RIGOR \cite{humayun_CVPR_2014_rigor}, and parametric min-loss \cite{Lee2015}. Some other methods utilized segmentation/superpixel grouping using, for instance, segment hierarchy \cite{Arbelaez_CVPR14, Rantalankila14, wangobject, Yanulevskaya14} or new distance measure \cite{xiao2015complexity}. In general, most of these methods can achieve good localization quality, but suffer from either poor computational efficiency or low DR during testing.

{\em (\rmnum{2}) Edge based algorithms:} Compared with segments and superpixels, edges are lightweight visual features in terms of computation. Currently most of the efficient proposal algorithms utilize edge related features. Zhang \etal \cite{zhang2011proposal} proposed a cascaded ranking SVM (CSVM) method to sample the proposals based on image gradients in a sliding-window manner, and later generalized the method into two-stage cascade SVMs in \cite{zhang2016object}. Cheng \etal \cite{BingObj2014} proposed the BING algorithm with binary features running at 300fps. Zhao \etal \cite{zhao2014cracking} showed that the success of BING is rather in combinatorial geometry and proposed a window sampling method accordingly. Zitnick and Doll{\'a}r \cite{zitnickedge} proposed the EdgeBoxes algorithm to fast generate proposals based on edges and contours while achieving good localization quality. Lu \etal \cite{Lu_ICCV2015} proposed a contour box algorithm to reject the object proposals without explicit closed contours. In addition, Qi \etal \cite{Qi2015} proposed a perceptual grouping framework that organizes image edges into meaningful structures, and tested this method for object proposal generation. Ghodrati \etal \cite{ghodrati2015deepproposal} proposed a DeepProposal method based on convolutional neural networks (CNN). Ren \etal \cite{girshick2015fast} proposed the region proposal network (RPN) to accelerate the fast R-CNN \cite{girshick2015fast} for object detection. In general, edge based algorithms are faster than segmentation based algorithm. Among them, BING is the fastest in the literature, but suffers from poor proposal localization quality seriously. 

{\em (\rmnum{3}) Proposal post-processing:} Several recent works focus on improving proposal quality with small amount of computational cost. For instance, He and Lau \cite{OOP_ICCV2015} proposed an oriented object proposal algorithm for better locating objects by estimating their orientations. Wang \etal \cite{wang2015improving} proposed using multi-thresholding straddling expansion (MTSE) to improve quality using superpixels.

A comprehensive comparison between some different object proposal algorithms can be found in \cite{Hosang2015Pami}. In contrast, our BING++ utilizes both edge and segmentation information sequentially to improve the proposal quality gradually, achieving better trade-off between proposal quality and computational efficiency in the literature.

\section{Understanding Object Proposals: A Probabilistic Perspective}\label{sec:understanding}
Let us consider the training (if any) and testing procedures separately for an object proposal algorithm. We denote as $\{\mathbf{x}_i\}_{i=1,\cdots,N}$ the training data with $N$ images, and as $\{\mathbf{s}_{ij}\}_{j=1,\cdots,N_i}, \forall i$ the ground-truth bounding box coordinates (\ie $\mathbf{s}_{ij}\in\mathbb{R}^4$) for $N_i$ objects in the $i$-th image, one box per object. Given a {\em proposal quality measure} $o$ and a corresponding {\em threshold} $\eta\geq0$, the training goal of a proposal algorithm is to determine a suitable {\em structured prediction} function (or mapping rule), $f^*$, to maximize the likelihood of correct detections (or equivalently) as follows:
\begin{align}\label{eqn:train}
f^* = \argmax_{f\in\mathcal{F}}\prod_{i=1}^N \prod_{j=1}^{N_i} P\left(\max_{\mathbf{y}\in f(\mathbf{x}_i)} o(\mathbf{y}, \mathbf{s}_{ij})\geq\eta \Big| \mathbf{x}_i, \mathbf{s}_{ij}\right), 
\end{align}
where $f:\mathbf{x}\rightarrow\mathcal{Y}\subseteq\mathbb{R}^4$ is the proposal generation function (or more generally an algorithm) from a feasible functional space $\mathcal{F}$ that extracts a collection of potential object regions $\mathcal{Y}$ as proposals from image $\mathbf{x}$, and $P$ denotes the conditional probability. Here function $o(\mathbf{y}, \mathbf{s}_{ij})$ measures the overlap (\eg using intersection-over-union (IoU)) between a proposal $\mathbf{y}$ and the ground-truth $\mathbf{s}_{ij}$. If this value is larger than $\eta$ (\eg $\eta=0.5$) we consider this proposal as a {\em correct} detection for the object. Then the likelihood in Eq.~\ref{eqn:train} essentially measures the joint probability of overall correct detections among the entire training data, which should be maximized by function $f^*$. Note that the functional space $\mathcal{F}$ could be restricted by certain (regularization) requirements such as number of proposals. For simplicity we do not explicitly show these requirements in the objective but assume that they are inherent in the definition of $\mathcal{F}$ implicitly. Similarly for the other functional spaces in the rest of the paper, we make the same assumption without explicit mention.

%

Recently researchers in computer vision have started to investigate the learning problems with such complicated quality measure \cite{nowozin2014optimal,ahmed2015optimizing}. For instance, Nowozin in \cite{nowozin2014optimal} studied the problem of making optimal decisions from probabilistic models with IoU scores and proposed a greedy algorithm to efficiently solve it. In contrast our learning problem in Eq. \ref{eqn:train} (and similarly in the related equations latter) can be potentially generalized to an arbitrary overlap measure with specific algorithms developed for generic object proposal generation (see Section \ref{sec:BING++}).


In test time we would like to generate proposals for possible objects in test images using the learned function $f^*$. As we see in Fig. \ref{fig:abo-stats}, there are very small portions of objects having best overlap (BO) scores less than $\eta=0.5$, indicating that the training procedure with $\eta=0.5$ works well. However, with the increase of threshold $\eta$, say to 0.7, as a new decision rule for correct detection, it is clear that for BING there will be a large portion of proposals which are considered as wrong detections. 

In order to solve this performance degradation problem, one possible solution is to retrain the models in Eq. \ref{eqn:train} directly with higher thresholds. This strategy, however, becomes more and more difficult with the increase of the threshold (see the experimental evaluation in \cite{zhang2011proposal, zhang2016object}), because in real data the parameter space for ground-truth bounding boxes of objects is so huge that localizing such windows accurately is extremely difficult with consideration of computational efficiency. \\

\noindent
\underline{\em Alternating optimization:} Another possible solution is to presume that the feasible functional space $\mathcal{F}$ has certain {\em structures}, so that we can shrink the searching space in $\mathcal{F}$, leading to much lower model complexity. Similar ideas have been explored in many different research areas, for instance, recently in developing efficient algorithms for training deep neural networks \cite{cheng2015exploration}, where circulant structures are used to simplify the fully-connected layers.

Particularly, in this paper we presume $\mathcal{F}$ as the composition of functional spaces as follows:
\begin{align}
\mathcal{F}\stackrel{\mbox{\scriptsize def}}{=}\mathcal{F}_1\times\mathcal{F}_2\times\dots\times\mathcal{F}_{M-1}\times\mathcal{F}_M,
\end{align}
where $\mathcal{F}_m, \forall m=1,\cdots,M$ denotes a feasible functional space. Accordingly we can view function $f^*$ as the composition of functions as follows:
\begin{align}\label{eqn:composition}
f^*\stackrel{\mbox{\scriptsize def}}{=}f_M^*\circ f_{M-1}^*\circ\dots\circ f_2^*\circ f_1^*, 
\end{align}
where $f_m^*, \forall m=1,\cdots,M$ denotes the $m$-th {\em atomic} localization function which is applied {\em sequentially} to generate proposals. A good example based on such methodology is \cite{wang2015improving}, where a superpixel merging technique was applied on top of existing proposal generators to improve localization quality. Considering the problem in Eq.~\ref{eqn:train}, such solutions from Eq. \ref{eqn:composition} are always {\em suboptimal} (with local optimality at $f_m^*$ in each functional space $\mathcal{F}_m$).

Intuitively this function composition in Eq. \ref{eqn:composition} suggests an alternating optimization routine to learn such atomic functions. That is, 
\begin{align}\label{eqn:alternating}
f_m^* = \argmax_{f_m\in\mathcal{F}_m}\prod_{i,j} P\left(\max_{\mathbf{y}\in f(\mathbf{x}_i)} o(\mathbf{y}, \mathbf{s}_{ij})\geq\eta \Big| \mathbf{x}_i, \mathbf{s}_{ij}\right), 
\end{align}
where $f\stackrel{\mbox{\scriptsize def}}{=}f^*_M\circ\dots\circ f_m \circ\dots\circ f^*_1, \forall m$ with the other fixed learned functions such as $f^*_1$ and $f^*_M$. Actually in our work \cite{zhang2013efficient} we have explored similar ideas and proposed a specific alternating optimization algorithm for learning these filters in cascade SVMs. \\

\noindent
\underline{\em Sequential optimization:} The main challenge of developing such alternating optimization algorithms for proposal generation lies in the fact that mathematical formulation of such ad hoc algorithms is extremely difficult to propose. Instead for simplicity in practice the atomic functions in these algorithms are usually learned/designed sequentially, such as \cite{wang2015improving}. Consequently this leads to the learning rule as 
\begin{align}\label{eqn:sequential}
f_m^* = \argmax_{f_m\in\mathcal{F}_m}\prod_{i,j} P\left(\max_{\mathbf{y}\in f_m(\bar{f}_m^*(\mathbf{x}_i))} o(\mathbf{y}, \mathbf{s}_{ij})\geq\eta \Big| \mathbf{x}_i, \mathbf{s}_{ij}\right), 
\end{align}
where $\bar{f}_m^*\stackrel{\mbox{\scriptsize def}}{=}f^*_{m-1}\circ\dots\circ f^*_1, \forall m$ is a fixed function that can be learned by solving Eq.~\ref{eqn:sequential} recursively for $m-1$ times. Obviously the maximum likelihood learned based on Eq.~\ref{eqn:sequential} is the lower bound of that based on Eq. \ref{eqn:alternating} and thus Eq.~\ref{eqn:train}. 

{\em In fact the training approach in our previous work \cite{zhang2011proposal,zhang2016object} as well as BING \cite{BingObj2014} is an exemplar that falls in this lower bound sequential maximization scheme.} First we learn a linear filter for each quantized scale/aspect-ratio in image space, and then on top of the filter responses we learn a second linear filter for final ranking purpose across different quantized scales/aspect-ratios. Here we take any feature from a window whose overlap with a ground-truth bounding box is larger than threshold $\eta$ as a positive instance for learning, otherwise as a negative one instead.


\section{BING++}\label{sec:BING++}
As we state before, 
due to the {\em fixed and unadaptive} quantization scheme, BING can only be considered as a {\em coarse} proposal generator. The goal of BING++ is to learn more functions {\em sequentially} on top of BING to refine its proposals efficiently and effectively. Note that the proposed method can also be applied on top of other proposal generation algorithms for refinement purpose. 
In order to achieve the {\em best} trade-off between localization quality and computational efficiency, we propose BING++ as our algorithm.

To parameterize Eq.~\ref{eqn:sequential} for sequential optimization, in this paper we choose to utilize Gaussian distributions with the 0/1-loss function to model the likelihood $P$ that maximizes DR given $\eta$. Other parameterizations may also be applied here, \eg maximizing MABO using Gaussian distributions with the least square loss and $\eta=1$, but how to select parameterization is beyond the scope of this paper. In summary, we parameterize $P$ in Eq.~\ref{eqn:sequential} as follows:
\begin{align}\label{eqn:P}
P\stackrel{\mbox{\scriptsize def}}{=}\exp\left[-\mathbf{1}_{\left\{\max_{\mathbf{y}\in f_m(\bar{f}_m^*(\mathbf{x}_i))} o(\mathbf{y}, \mathbf{s}_{ij}) < \eta \right\}}\right],
\end{align}
where $\bar{f}_m^*\stackrel{\mbox{\scriptsize def}}{=}f^*_{m-1}\circ\dots\circ f^*_1\circ f^*_{B}, \forall m$, $f^*_B$ denotes the learned sequential functions by BING after training (\ie $f^*_B(\mathbf{x}_i), \forall i$ represents the output proposals by BING for image $\mathbf{x}_i$), $\mathbf{1}_{\{\cdot\}}$ denotes the binary indicator function measuring the localization quality of proposals generated by function $f_m$, and it returns 1 if the condition is true, otherwise 0. Then by taking the log operation on the right hand side of Eq.~\ref{eqn:sequential}, we can write the {\em log-likelihood} of Eq.~\ref{eqn:sequential} as follows:
\begin{align}\label{eqn:0/1}
\hspace{-3mm} f^*_m=\argmin_{f_m\in\mathcal{F}_m}\sum_{i,j}\mathbf{1}_{\left\{\max_{\mathbf{y}\in f_m(\bar{f}_m^*(\mathbf{x}_i))} o(\mathbf{y}, \mathbf{s}_{ij}) < \eta \right\}}.
\end{align}

Recall that the goal of our BING++ is to generate object proposals accurately as well as efficiently (\ie fast running speed). To achieve this goal, we propose a specific algorithm, {\em RecursiveBox}, with consideration of both edge and segment information for refining the proposals generated by BING. RecursiveBox is developed towards optimizing Eq.~\ref{eqn:0/1} directly, which is non-trivial, because Eq.~\ref{eqn:0/1} is highly non-convex. Even if we relax the 0/1-loss function to its convex surrogate loss (\eg hinge-loss), as traditional methods such as SVMs, the resulting optimization problem will be still highly non-convex due to the non-convexity of both $\max$ operator and overlap measure function $o$.

To solve Eq.~\ref{eqn:0/1}, we propose a very efficient algorithm to search for the {\em approximate} solution in a {\em quantized} parameter space. We deliberately design function $f_m$ for different image cues (\ie edges and segments in this paper particularly), and quantize the corresponding parameter space into a finite number of disjoint subspaces, represented by the centroid parameter of each subspace (similar to cluster centers in KMeans). Then using each representative parameter, we can simply compute the loss in Eq. \ref{eqn:0/1}. By collecting all the losses we can select a representative parameter as the approximate solution which leads to a minimum loss. 

Intuitively our RecursiveBox works as follows:  At time $t$, we estimate a new bounding box for objects based on both image cues and the location/scale of the current bounding box. If both bounding boxes share a large overlap, we consider the new box as a good estimator/proposal, because it is stabilizing. Otherwise, we update the current bounding box with the new one for time $t+1$. This {\em deterministic} rule is repeated over time until some termination criterion (\eg number of iterations) is satisfied. In each update we equivalently learn a function as $f_m$ in Eq. \ref{eqn:0/1} that can be applied sequentially. Meanwhile we preserve all the objectness scores associated with the initial bounding boxes. 

In the following sections, we first explain how to refine proposals using edge information in Section \ref{ssec:edge}, which is extremely computationally efficient. Then we introduce segment information as the other cue for refinement in Section \ref{ssec:seg}, which results in higher computational burden but generates proposals with better localization quality. Finally we integrate both image cues together and propose our BING++ algorithm in Section \ref{ssec:bing++}.



\subsection{Edge-based Refinement}\label{ssec:edge}
Object proposal generation is about precisely capturing object boundaries, regardless of the pixels inside objects. We observe that good bounding boxes are those that tightly cover object boundaries. Edges as computationally efficient (compared to superpixels or segments) and indicative features to object boundaries are usually utilized to approximate object boundaries \cite{arbelaez2011contour, dollar2006supervised, martin2004learning}. Indeed most of the efficient proposal algorithms utilize edge related features. 
If the boundary of a proposal does not intersect with any edge point, we can speculate that either this proposal does not cover any object, or it is too loose. Similarly, if the proposal intersects with too many edge points, we expect that it has not yet reached the object boundary. In all of these cases, we would modify the proposals by pushing them towards the object boundaries. Therefore, {\em we propose utilizing the nearest edge points to proposal boundaries as {\bf fast yet weak} indicators of object boundaries to refine the proposals}.

Given an image, we denote $\mathbf{r}(t)\in\mathbb{R}^4 (t\geq0)$ as the predicted bounding box at time $t$, 
$\mathcal{A}(\mathbf{r}(t))\subseteq\mathbb{R}^2$ as the set of pixel locations in the area covered by $\mathbf{r}(t)$, and $\mathcal{B}\subseteq\mathbb{R}^2$ as the {\em constant} edge map of the image. 
We then generate a pixel location set, $\mathcal{C}(\mathbf{r}(t))$, for $\mathcal{A}(\mathbf{r}(t))$ by looking for the nearest neighbors in $\mathcal{B}$. That is,
\begin{align}\label{eqn:C}
\mathcal{C}(\mathbf{r}(t))=\left\{\mathbf{q}\Big|\argmin_{\mathbf{q}\in\mathcal{B}}d(\mathbf{p}, \mathbf{q}), \forall \mathbf{p}\in\mathcal{A}(\mathbf{r}(t))\right\}\subseteq\mathbb{R}^2,
\end{align}
where $d(\cdot,\cdot)$ denotes a distance function. For some special distance metric such as Euclidean distance, $\mathcal{C}(\mathbf{r}(t))$ can be efficiently computed using distance transform \cite{journals/toc/FelzenszwalbH12}\footnote{A good tutorial on how to locate the nearest edge point for each pixel using distance transform can be found at \url{http://www.vlfeat.org/overview/imdisttf.html}.}.

Based on $\mathcal{C}(\mathbf{r}(t))$, we define the predicted bounding box, $\mathbf{r}(t+1)$, at time $t+1$ as follows:
\begin{align}\label{eqn:r(t+1)}
\mathbf{r}(t+1) = (1-\gamma)\mathbf{r}(t) + \gamma \left[\min_{\mathbf{q}\in\mathcal{C}(\mathbf{r}(t))}\mathbf{q};\max_{\mathbf{q}\in\mathcal{C}(\mathbf{r}(t))}\mathbf{q}\right],
\end{align}
where $\min, \max$ are entry-wise minimum and maximum operators, $[\cdot;\cdot]$ denotes the vector concatenating operator, and $0\leq\gamma\leq 1$ is a trade-off parameter. The basic idea of Eq. \ref{eqn:r(t+1)} is to generate a new box by linearly integrating the two boxes. 

By substituting Eq. \ref{eqn:r(t+1)} into Eq. \ref{eqn:0/1}, we have the following optimization problem for refining boxes based on edges:
\begin{align}\label{eqn:edge_opt}
\min_{0\leq\gamma\leq1}\sum_{i,j}\mathbf{1}_{\left\{\max_k o(\mathbf{r}_{ik}(t+1), \mathbf{s}_{ij}) < \eta \right\}},
\end{align}
where $\mathbf{r}_{ik}(t+1), \forall i,\forall k$ denotes the $k$-th new bounding box generated from the $k$-th current bounding box $\mathbf{r}_{ik}(t)$ in the $i$-th image. \\

\begin{figure}[t]
\begin{minipage}[b]{0.48\linewidth}
 \begin{center}
 \centerline{\includegraphics[width=\columnwidth]{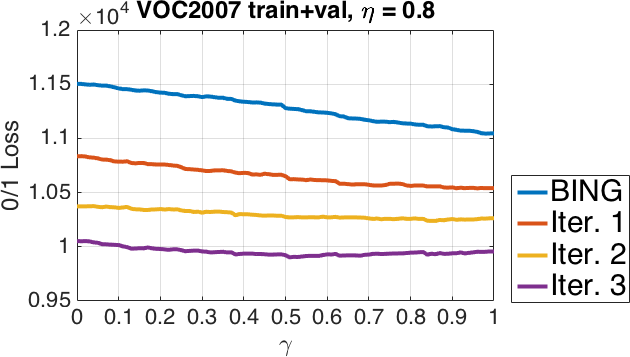}}
 \end{center}
\end{minipage}
\begin{minipage}[b]{0.48\linewidth}
\begin{center}
\centerline{\includegraphics[width=\columnwidth]{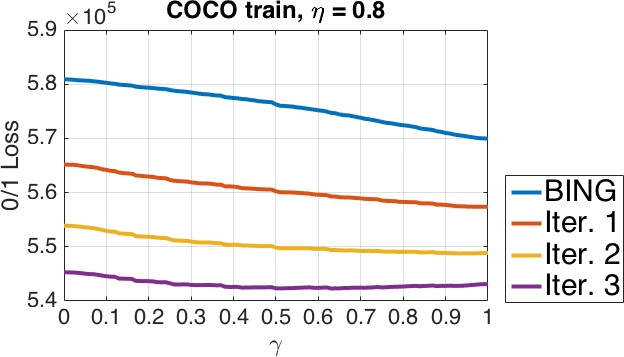}}
\end{center} 
\end{minipage}
\vspace{-5mm}
\caption{\footnotesize Statistical behavior comparison when optimizing Eq. \ref{eqn:edge_opt} on VOC2007 train+val (left) and COCO training (right) datasets using $\eta=0.8$ as the threshold for measuring high-quality proposals. 
}\label{fig:gamma}
\vspace{-3mm}
\end{figure}


\noindent
\underline{\em Optimization:} 
Since the parameter $\gamma$ in Eq. \ref{eqn:edge_opt} is a scalar, we simply enumerate all possible {\em quantized} values as approximate solutions to accelerate the learning. Specifically, we quantize $\gamma$ from 0 to 1, step by 0.01, and perform {\em greedy search} over iterations by computing the loss in Eq. \ref{eqn:edge_opt} based on each quantized value. Fig.~\ref{fig:gamma} illustrates the loss statistics with overlap threshold $\eta=0.8$ (for $\eta>0.8$ we have similar observations) because we would like to generate high-quality proposals. Note that we learn one $\gamma$ in each iteration which results in a sequential function that can be applied in Eq. \ref{eqn:0/1}. To plot these curves we set the parameter $\gamma$ for the next iteration as the one achieving the minimum loss in the current iteration. In each iteration the inputs for the function are the outputs from the previous iteration (or BING proposals as initialization). We repeat the same procedure to compute the overall loss in Eq. \ref{eqn:edge_opt} on both VOC2007 train+val and COCO training datasets to verify whether we can learn the parameter with good generalization. Indeed Fig.~\ref{fig:gamma} has demonstrated strong similarities between the statistical behaviors on both datasets, suggesting that we can generalize the learned $\gamma$ values across different datasets. With increasing number of iterations the loss curves become flat on both figures, indicating the convergence of our algorithm empirically. \\


\begin{algorithm}[t]\footnotesize
\SetAlgoLined
\SetKwInOut{Input}{Input}\SetKwInOut{Output}{Output}
\Input{edge-based distance transform map $\mathcal{C}$, proposals $\mathcal{R}$, overlap threshold $\epsilon\geq0$}
\Output{improved proposals $\Omega$}
\BlankLine
$\Omega\leftarrow\emptyset$;\\
\ForEach{$\mathbf{r}_i(0)\in\mathcal{R}$}
{
\For{$t = 0$ \KwTo $T-1$}
{
$\mathbf{r}_i(t+1) \leftarrow \left[\min_{\mathbf{q}\in\mathcal{C}(\mathbf{r}_i(t))}\mathbf{q};\max_{\mathbf{q}\in\mathcal{C}(\mathbf{r}_i(t))}\mathbf{q}\right]$;\\
\lIf{$o(\mathbf{r}_i(t),\mathbf{r}_i(t+1))\geq\epsilon$}{break}
}
$\Omega\leftarrow\Omega\bigcup\mathbf{r}_i(t+1)$
}
\Return $\Omega$
\caption{\small EdgeRecursiveBox algorithm}\label{alg:edge}
\end{algorithm}

\noindent
\underline{\em Implementation:} 
We list our EdgeRecursiveBox algorithm in Alg. \ref{alg:edge} with computational complexity of $O(|\mathcal{R}|\cdot T)$, roughly speaking, where $|\mathcal{R}|$ denotes the number of input proposals in set $\mathcal{R}$, and $T$ denotes the number of iterations. We utilize canny edge detection to create edge maps $\mathcal{B}$ to approximate object boundaries. Ideally accurately detecting object boundaries is very desirable yet challenging due to complex imaging factors and semantic ambiguity, and many existing works in the literature such as \cite{arbelaez2011contour,dollar2006supervised,martin2004learning} followed similar ideas. Better boundary detection such as structured edges \cite{dollar2013structured} may improve the proposal quality at the cost of longer computational time. We utilize Euclidean distance for function $d$ in Eq. \ref{eqn:C} and thus employ distance transform to compute nearest edge maps $\mathcal{C}$. By taking into account both generalization and computational complexity and based on the observation from Fig.~\ref{fig:gamma}, we specifically set $\gamma=1, T=3$. When $\gamma=1$ for each iteration, the statistical behaviors on both datasets are almost identical to those in Fig. \ref{fig:gamma}. Parameter $\epsilon$ is predefined to determine whether the procedure of updating a bounding box converges, and empirically we set $\epsilon=0.95$. In our experiments we utilize these default values for all the datasets.

To accelerate the computation in EdgeRecursiveBox, we resize images into $1/3\times1/3=1/9$ of their original sizes. The reasons for doing are based on the observations: (1) Distance transform is relatively time-consuming, whose complexity is linearly propositional to image size. (2) Proposals hardly localize small objects correctly in the original images. In other words, discarding the detections of small objects has little effect on localization quality measure (\ie ABO and MABO). Therefore, our image resizing operation leads to marginal performance degradation but significant speed-up (see our comparison in Section \ref{ssec:derivatives}).

\begin{figure}[t]
\begin{minipage}[b]{0.49\linewidth}
 \begin{center}
 \centerline{\includegraphics[width=.7\columnwidth]{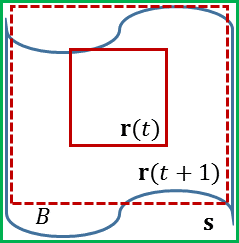}}
 \centerline{\footnotesize{(a)}}
 \end{center}
\end{minipage}
\begin{minipage}[b]{0.49\linewidth}
\begin{center}
\centerline{\includegraphics[width=.7\columnwidth]{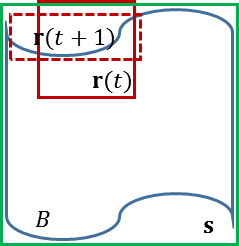}}
 \centerline{\footnotesize{(b)}}
\end{center} 
\end{minipage}
\vspace{-5mm}
\caption{\footnotesize{Illustration of updating current red solid bounding box $\mathbf{r}(t)$ to next red dashed bounding box $\mathbf{r}(t+1)$. Our estimation for the ground-truth bounding box $\mathbf{s}$ based on $\mathcal{C}(\mathbf{r}(t))$ succeeds in (a) where the pixels in $\mathcal{C}(\mathbf{r}(t))$ spread well, but fails in (b) where the pixels in $\mathcal{C}(\mathbf{r}(t))$ concentrate on few boundary fragments.}}\label{fig:sampling}
\vspace{-3mm}
\end{figure}

\subsection{Segmentation-based Refinement}\label{ssec:seg}
Though edges are fast to compute for approximating object boundaries, their stability is very limit. Missing boundary fragments often occur with usage of edge detection, as there is no strong contrast at such places. This will cause serious trouble to our EdgeRecursiveBox algorithm because object boundaries cannot be inferred for precise coverage. Also there is no guarantee that our estimator $\mathbf{r}(t+1)$ will approach the ground-truth bounding box eventually. Fig. \ref{fig:sampling} illustrates two simple cases where our estimation succeeds and fails, respectively, given sufficient edge information. In (a) the current bounding box $\mathbf{r}(t)$ is surrounded by the edge points in $\mathcal{B}$, implying that the pixels in $\mathcal{C}(\mathbf{r}(t))$ are sufficiently well-spread. This improves the estimate for the ground-truth bounding box $\mathbf{s}$. In (b), $\mathbf{r}(t)$ intersects with the edge points. This leads to a situation where $\mathcal{C}(\mathbf{r}(t))$ is determined by a small fraction of $\mathcal{B}$. In practice, there may be scenarios where the correct detections in BING could be updated to wrong bounding boxes.

To deal with these challenges in EdgeRecursiveBox, we further consider segments as a second type of useful cues which are usually more indicative of boundaries (even the coverage of objects) at the cost of more computational time. 
Meanwhile, bounding boxes always have overlaps with certain segments, making it possible to improve the boxes based on the boundaries of segments. Therefore, {\em we propose utilizing segments as relatively {\bf slow yet strong} indicators of object boundaries to refine our proposals as well}.

Given a bounding box $\mathbf{r}(t)$ at time $t$, the training goal of our segmentation based refinement is to learn a function (or algorithm in general) by combining segments with $\mathbf{r}(t)$ to generate a new box $\mathbf{r}(t+1)$ so that Eq.~\ref{eqn:0/1} is  minimized. In test time, we apply the same function to all the bounding boxes at time $t$ for updating purpose. Letting $\mathcal{S}'$ denote the set of selected segments for updating $\mathbf{r}(t)$, we formally define the new box $\mathbf{r}(t+1)$ as follows:
\begin{align}\label{eqn:seg_box}
\hspace{-3mm}\mathbf{r}(t+1)=\left[\min_{\mathbf{q}\in\mathcal{A}(\mathbf{r}(t))\bigcup\mathcal{A}(\mathcal{S}')}\mathbf{q}; \max_{\mathbf{q}\in\mathcal{A}(\mathbf{r}(t))\bigcup\mathcal{A}(\mathcal{S}')}\mathbf{q}\right],
\end{align}
where $\mathcal{A}(\mathcal{S}')\subseteq\mathbb{R}^2$ denotes the set of pixel locations covered by the segments in $\mathcal{S}'$. 

In order to update bounding boxes using Eq. \ref{eqn:seg_box}, intuitively we need to find a way to select relavent segments for $\mathcal{S}'$. By considering this as well as minimizing Eq. \ref{eqn:0/1} we have the following general optimization problem for refining boxes based on segments:
\begin{align}\label{eqn:segment_opt}
\min_{g\in\mathcal{G}} & \sum_{i,j}\mathbf{1}_{\left\{\max_k o(\bar{\mathbf{r}}_{ik}, \mathbf{s}_{ij}) < \eta \right\}}, \\
\mbox{s.t.} & \bar{\mathbf{r}}_{ik}=\left[\min_{\mathbf{q}\in\mathcal{A}(\mathbf{r}_{ik}(t))\bigcup\mathcal{A}(\mathcal{S}_{ik}')}\mathbf{q}; \max_{\mathbf{q}\in\mathcal{A}(\mathbf{r}_{ik}(t))\bigcup\mathcal{A}(\mathcal{S}_{ik}')}\mathbf{q}\right], \label{eqn:seg_opt_1}\\
& \mathcal{S}_{ik}' = g(\mathbf{r}_{ik}(t), \mathcal{S}_i), \; \forall i, \forall k, \nonumber
\end{align}
where $\bar{\mathbf{r}}_{ik}, \forall i, \forall k$ denotes the $k$-th new bounding box generated from the $k$-th current bounding box $\mathbf{r}_{ik}(t)$ in the $i$-th image, $\mathcal{S}_i, \forall i$ denotes the segment set in the $i$-th image, $g$ denotes a segment selection function to generate the selected segment set $\mathcal{S}_{ik}'$ for $\mathbf{r}_{ik}(t)$, and $\mathcal{G}$ denotes its feasible functional space. \\

\noindent
\underline{\em Optimization:}
The problem in Eq. \ref{eqn:segment_opt}, in general, can be considered as a combinatorial optimization problem \cite{cook1998combinatorial}. However, solving Eq. \ref{eqn:segment_opt} with an exponential number (\wrt segments) of potential new boxes for each current box is extremely difficult, especially when considering computational efficiency. 

\begin{figure}[t]
\begin{minipage}[b]{0.5\linewidth}
	\begin{center}
		\centerline{\includegraphics[width=1.1\columnwidth]{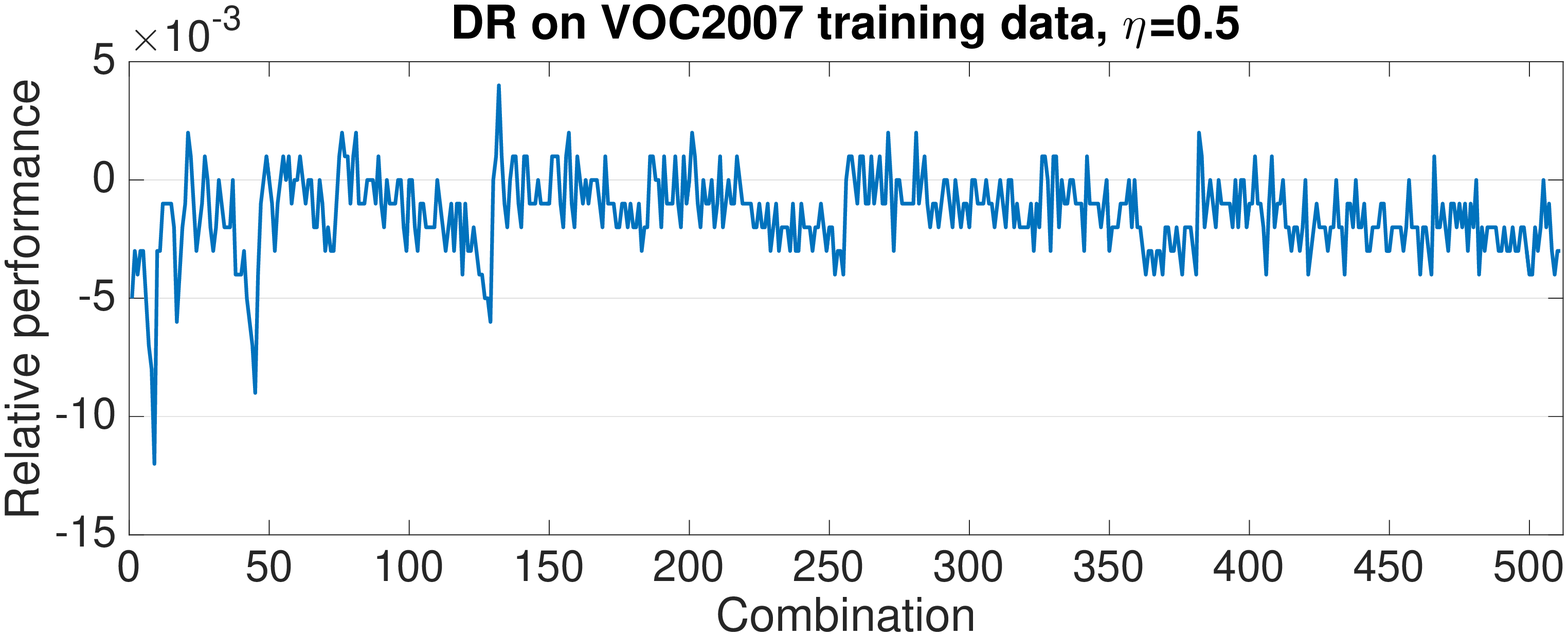}}
	\end{center}
\end{minipage}
\begin{minipage}[b]{0.5\linewidth}
	\begin{center}
		\centerline{\includegraphics[width=1.1\columnwidth]{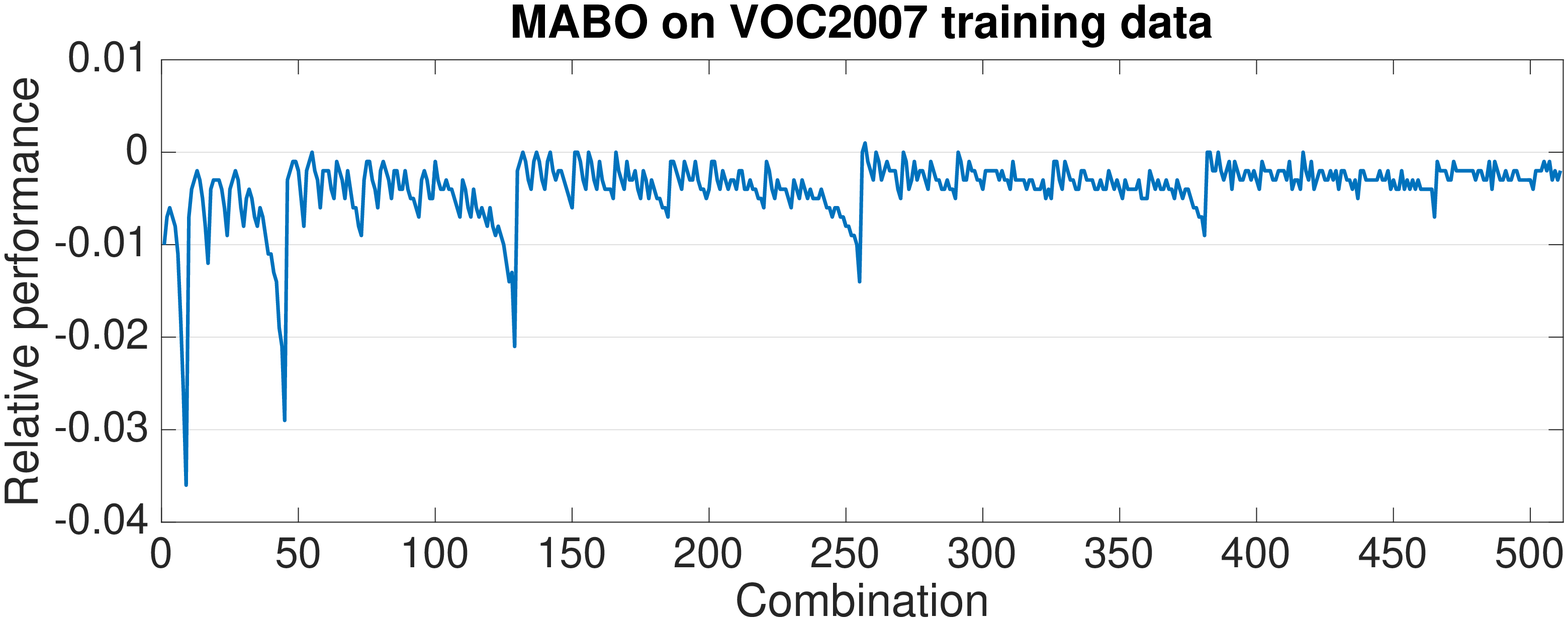}}
	\end{center} 
\end{minipage}	
\begin{minipage}[b]{0.5\linewidth}
 \begin{center}
 \centerline{\includegraphics[width=1.1\columnwidth]{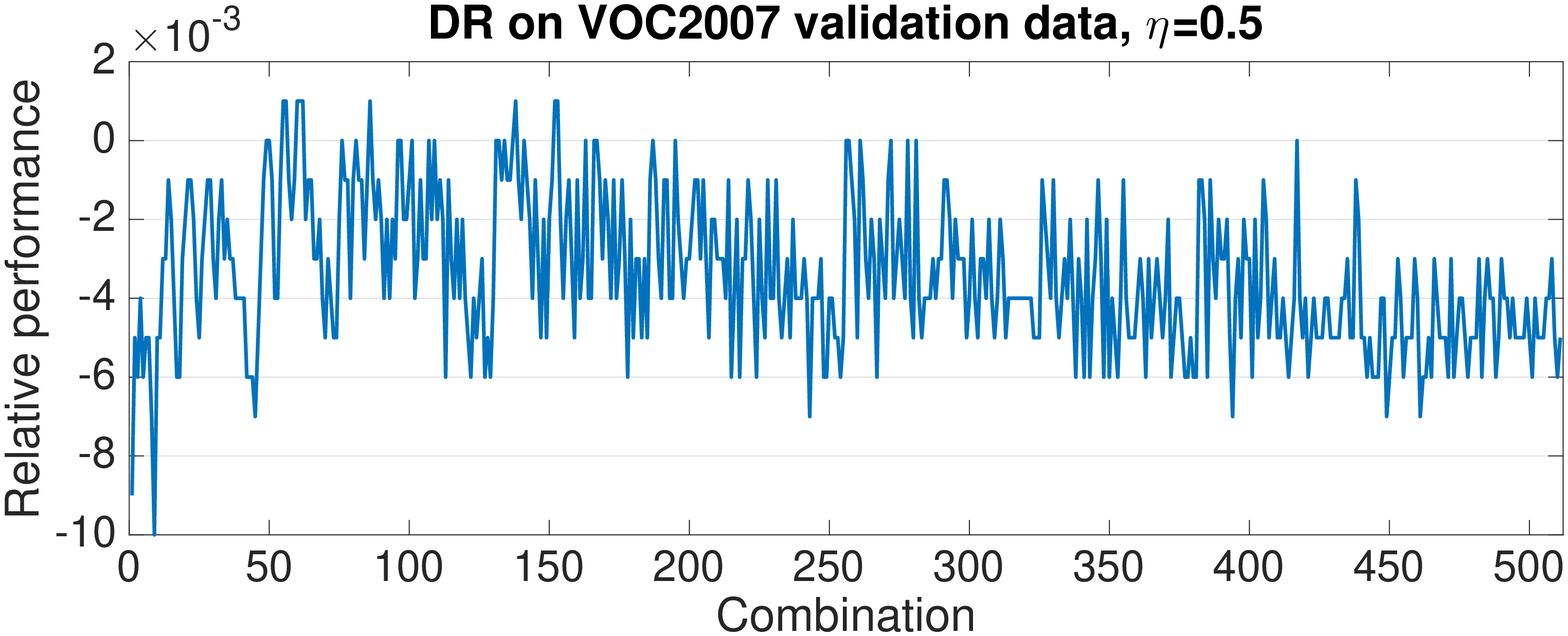}}
 \end{center}
\end{minipage}
\begin{minipage}[b]{0.5\linewidth}
\begin{center}
\centerline{\includegraphics[width=1.1\columnwidth]{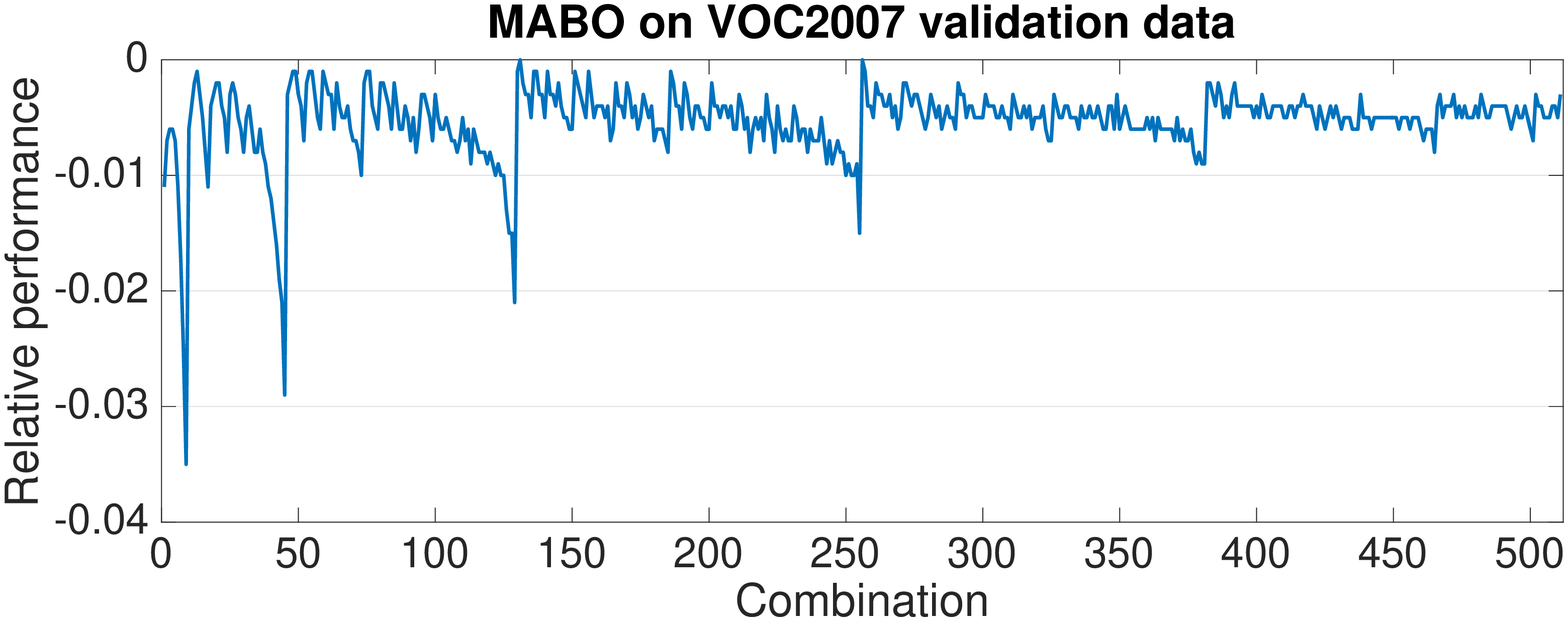}}
\end{center} 
\end{minipage}
\begin{minipage}[b]{0.5\linewidth}
 \begin{center}
 \centerline{\includegraphics[width=1.1\columnwidth]{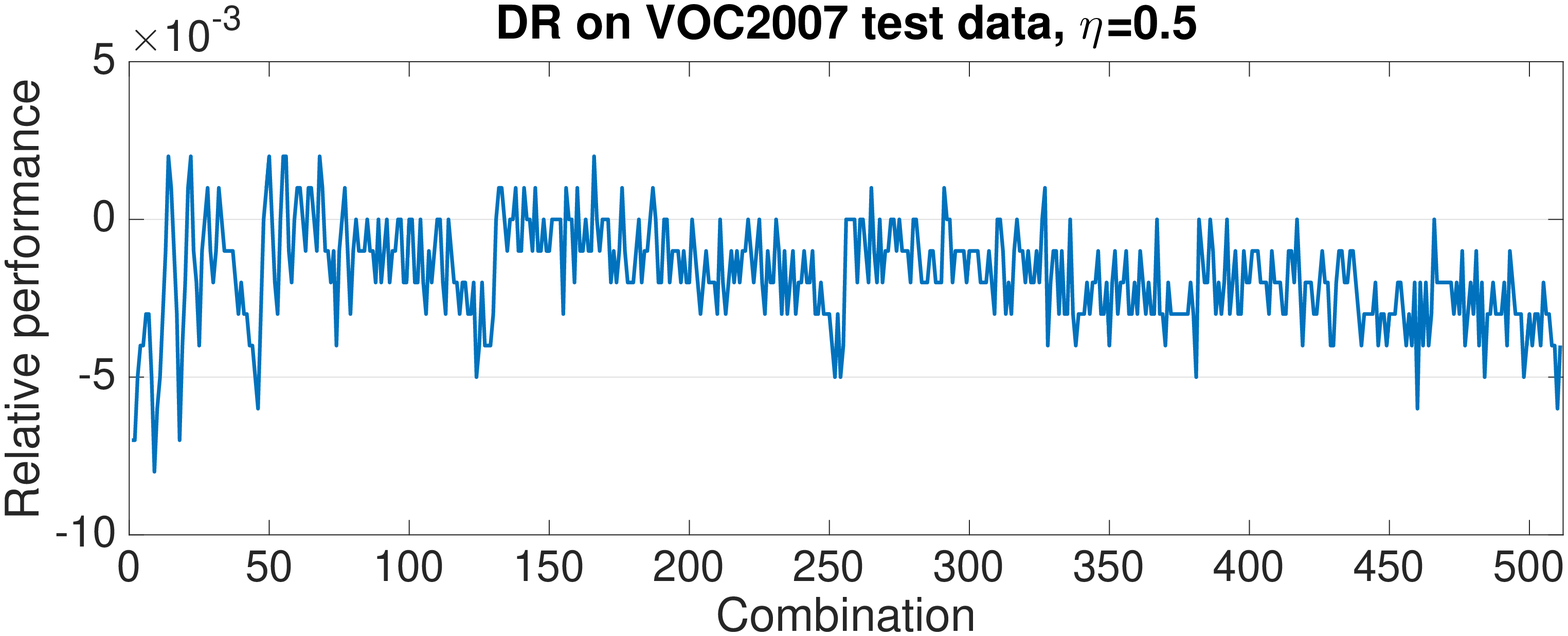}}
 \end{center}
\end{minipage}
\begin{minipage}[b]{0.5\linewidth}
\begin{center}
\centerline{\includegraphics[width=1.1\columnwidth]{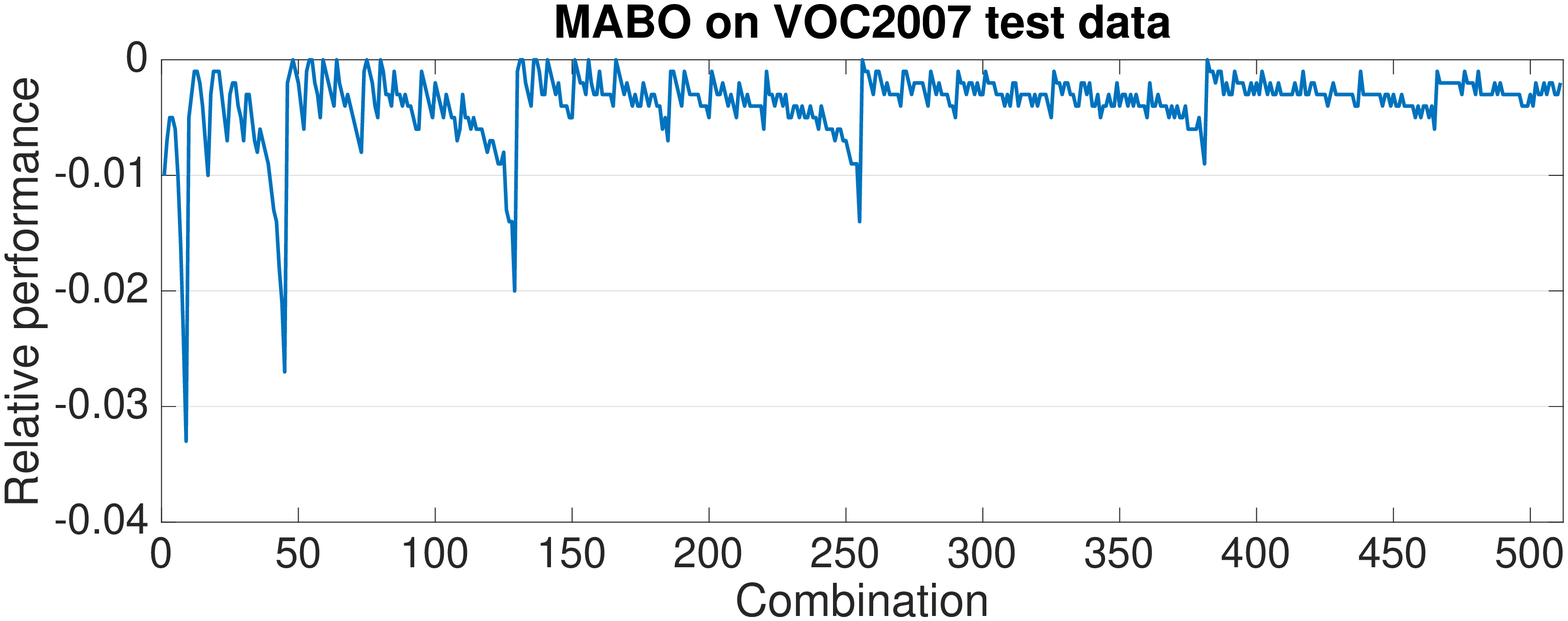}}
\end{center} 
\end{minipage}
\vspace{-8mm}
\caption{\footnotesize Statistical behavior comparison on DR/MABO \vs $\Delta$ using VOC2007 training ({\bf top}), validation ({\bf middle}), and test ({\bf bottom}) datasets, respectively, by minimizing Eq. \ref{eqn:segment_opt}. Here x-axis shows the indexes of all possible combinations in $\Delta$, and y-axis shows the performance improvement \wrt that with the combination $\{0.1, 0.2, 0.3, 0.4, 0.5\}$ used in \cite{wang2015improving}. 
}\label{fig:seg_stat}
\vspace{-3mm}
\end{figure}

Instead here we utilize our {\em parameter space quantization} mechanism again to reduce the complexity of our combinatorial optimization problem deliberately so that we can search for corresponding new boxes efficiently as approximate solutions. 
We prefer the complexity no higher than {\em linear} per box in terms of number of segments. 
To do so, we propose grouping segments in each image into several {\em subsets} (with overlaps) for each current bounding box and then taking one bounding box per subset which tightly covers the corresponding segments as a new bounding box. In such way, we can approximate the optimal $g$ based on the overlap function and a set of thresholds, and further rewrite Eq. \ref{eqn:segment_opt} as follows: $\forall i, \forall k, \forall\delta\in\Delta,$
\begin{align}\label{eqn:seg_opt}
\mathcal{S}_{ik}' = g(\mathbf{r}_{ik}(t), \mathcal{S}_i) = \left\{s_l|\bar{o}(s_l, \mathbf{r}_{ik}(t)) \geq \delta, \forall s_l\in\mathcal{S}_i\right\}, \; 
\end{align}
where $\bar{\mathbf{r}}_{ik}^{(\delta)}, \forall i, \forall k$ denotes a new bounding box at time $t+1$ parametrized by threshold $\delta (0\leq\delta\leq1)$ in the threshold set $\Delta$, $s_l\in\mathcal{S}_i, \forall l$ denotes the $l$-th segment in the $i$-th image, and $\bar{o}(s_l, \mathbf{r}_{ik}(t))$ denotes an overlap scoring function between segment $s_l$ and bounding box $\mathbf{r}_{ik}(t)$, defined as follows:
\begin{align}\label{eqn:seg_overlap}
\bar{o}(s_l, \mathbf{r}_{ik}(t)) = \frac{|\mathcal{A}(s_l)\cap\mathcal{A}(\mathbf{r}_{ik}(t))|}{|\mathcal{A}(s_l)|}\in[0,1], \forall l, \forall i, \forall k,
\end{align}  
where $\mathcal{A}(s_l)\subseteq\mathbb{R}^2, \forall l$ denotes the set of pixel locations covered by the segment $s_l$, $\cap$ denotes the set intersection operator, and $|\cdot|$ denotes the set cardinality.

Now our goal is to learn $\Delta$. To do so, we optimize Eq.~\ref{eqn:segment_opt} with Eq.~\ref{eqn:seg_opt} using nine {\em quantization values} for $\delta\in\Delta$, that is, from 0.1 to 0.9, step by 0.1. Then the total number of possible combinations for constructing $\Delta$ is $2^9-1=511$, which are easy to be tested on the data by minimizing Eq.~\ref{eqn:segment_opt}. For instance, in order to compare the statistical behavior on localization quality (\ie DR and MABO) using different combinations for $\Delta$, we show our results on VOC2007 training, validation and test datasets in Fig.~\ref{fig:seg_stat}. We can observe clearly that the statistical behaviors on both datasets are very similar, indicating that {\em the learned threshold set $\Delta$ may have good generalization across different datasets.} \\


\noindent
\underline{\em Implementation:}
We show our SegmentRecursiveBox algorithm for bounding box refinement in Alg. \ref{alg:seg}. The computational complexity of Alg. \ref{alg:seg} is $O(|\mathcal{R}|\cdot|\Delta|\cdot|\mathcal{S}|)$, roughly speaking, where $|\mathcal{R}|, |\Delta|, |\mathcal{S}|$ denote the numbers of bounding boxes, thresholds, and segments in images, respectively. 

We notice that our algorithm shares many similarities with \cite{wang2015improving} in terms of implementation. However, one of the key differences between our method and \cite{wang2015improving} is the {\em perspective of determining the thresholds in $\Delta$.} In \cite{wang2015improving} the parameters are fixed as $[0.1, 0.2, 0.3, 0.4, 0.5]$ by fitting the distribution of superpixel tightness with equal importance. As we see in Fig. \ref{fig:seg_stat}, based on our objective the parameter combination in \cite{wang2015improving} is not the best, and many other parameter combinations can achieve very similar performance as \cite{wang2015improving}. In contrast our method {\em learns} these parameters discriminatively by (approximately) minimizing 0/1-loss in Eq.~\ref{eqn:segment_opt}. {\em Empirically we set $\Delta = \{0.1, 0.3, 0.6\}$ by default for all the experiments}. 

\begin{algorithm}[t]\footnotesize
	\SetAlgoLined
	\SetKwInOut{Input}{Input}\SetKwInOut{Output}{Output}
	\Input{segment set $\mathcal{S}$, proposals $\mathcal{R}$, multiple thresholds $\Delta$}
	\Output{improved proposals $\Omega$}
	\BlankLine
	$\Omega\leftarrow\emptyset$;\\
	\ForEach{$\mathbf{r}_i(t)\in\mathcal{R}$}
	{
		\ForEach{$s_l\in\mathcal{S}$}
		{
			Compute $o(s_l,\mathbf{r}_i(t))$ based on Eq. \ref{eqn:seg_overlap};\\
			\ForEach{$\delta_j\in\Delta$}
			{
				\If{$o(s_l,\mathbf{r}_i(t))\geq\delta_j$}
				{
					Update $\mathbf{r}_i^{(\delta_j)}(t+1)$ based on Eq. \ref{eqn:seg_opt_1} and Eq. \ref{eqn:seg_opt};										
				}
			}
		}
		\ForEach{$\delta_j\in\Delta$}
		{
			$\Omega\leftarrow\Omega\bigcup\{\mathbf{r}_i^{(\delta_j)}(t+1)\}$;
		}
	}
	\Return $\Omega$
	\caption{\small SegmentRecursiveBox algorithm}\label{alg:seg}
\end{algorithm}

We employ \cite{Felzenszwalb:2004:EGI:981793.981796} to generate segments as it can achieve good performance as well as computational efficiency. From our experiments we find that the segmentation approaches which generate segments along gradients (usually leading to larger segments) contribute significantly to the success of segmentation based box refinement algorithms such as \cite{wang2015improving}. For comparison we replace \cite{Felzenszwalb:2004:EGI:981793.981796} with mean-shift \cite{comaniciu2002mean} and regenerate proposals on VOC2007 test dataset. We observe slight performance degradation in such way by 1.9\% and 3.8\% in terms of DR and MABO, respectively. When we utilize mean-shift with a superpixel combination post-process, same as the function $meanShiftSegmentation$ in openCV, our performance degrades only by 0.8\% and 0.7\% for DR and MABO, respectively.

As the computational complexity of \cite{Felzenszwalb:2004:EGI:981793.981796} scales linearly with the number of input nodes in the graphs, in general, we decide to utilize ``dense sampling'' to generate superpixels from images as input nodes, rather than utilizing pixels directly as did in \cite{wang2015improving}, to further accelerate the computation. Specifically we resize each image to $360\times400$ pixels, and take every $4\times4$ pixels as a cell without overlap, leading to $90\times100$ cells to form a grid per image. We then take each cell as a superpixel and feed all the cells to \cite{Felzenszwalb:2004:EGI:981793.981796}. In such way we observe significant speed-up with slight performance degradation (see our comparison in Section~\ref{ssec:derivatives}). We notice that different superpixel generation algorithms do have significant impact on the trade-off between proposal quality and computational efficiency. For instance, if utilizing gSLICr \cite{gSLICr_2015} to generate superpixels, we can process an image with a NVIDIA GeForce GTX 980 using about 5ms (slower than BING++'s 3ms), but achieve about 95.3\% and 79.2\%  (better than BING++'s 93.7\% and 77.5\%) in terms of DR and MABO, respectively, on VOC2007 using $\eta=0.5$. In this paper, however, we are not pursuing GPU acceleration in order to compare our BING++ with other algorithms in the literature fairly. 

\subsection{BING++ Algorithm}\label{ssec:bing++}
Overall, our proposed BING++ algorithm in Alg. \ref{alg:bing++} is essentially a sequential combination of BING, edge-based refinement as one ``+'', and segmentation-based refinement as the other ``+''. We set $\rho = 0.85$ for NMS by default. BING++ retains BING's DR performance while improving MABO with little degradation in computational time. The computational complexity of BING++ is dominated linearly by both image resolution and number of proposals.

We also test the other possibility of refining BING proposals using segments first and then edges. Compared with BING++ we observe performance degradation by 0.5\% and 2.1\% on VOC2007 test dataset, and 2.3\% and 2.9\% on COCO validation dataset, respectively, in terms of DR and MABO with 1,000 proposals and $\eta=0.5$. This is understandable, because edge based refinement is too loose, leading to large deviation from the true object locations for some good proposals generated by segmentation based refinement.

\begin{algorithm}[t]\footnotesize
\SetAlgoLined
\SetKwInOut{Input}{Input}\SetKwInOut{Output}{Output}
\Input{an input image $I$, edge based overlap threshold $\epsilon\geq0$, multiple thresholds $\Delta$, NMS parameter $\rho\geq0$}
\Output{generic object proposals $\Omega$}
\BlankLine

\tcp{BING proposals}
$\mathcal{R}\leftarrow\mbox{BING}(I)$;\\

\tcp{edge based refinement (\ie E-BING)}
$\mathcal{B}\leftarrow\mbox{CannyEdgeDetection}(I)$; $\mathcal{C}\leftarrow\mbox{DistanceTransform}(\mathcal{B})$;\\
$\Omega\leftarrow\mbox{EdgeRecursiveBox}(\mathcal{C}, \mathcal{R}, \epsilon)$;\\
\vfill
\tcp{segmentation based refinement (\ie S-BING)}
$\mathcal{S}\leftarrow\mbox{Segmentation}(I)$;
$\Omega\leftarrow\mbox{SegmentRecursiveBox}(\mathcal{S}, \Omega, \Delta)$;\\
$\Omega\leftarrow\mbox{NMS}(\Omega, \rho)$; \\
\Return $\Omega$
\caption{\small Test-time BING++ for object proposals}\label{alg:bing++}
\end{algorithm}


\section{Experiments}\label{sec:exp}
We conduct comprehensive experiments to demonstrate that BING++ is extremely efficient as well as capable of generating high quality object proposals. 

We test our method on the PASCAL VOC2007 \cite{pascal-voc-2007} and Microsoft COCO \cite{lin2014microsoft} datasets. VOC2007 contains 20 object categories, and consists of 9,963 natural images with object labels and their corresponding ground-truth bounding boxes released for training, validation and test sets. There are 5,011 images in the training and validation datasets, in total, and 4,952 images in test dataset. We use the training dataset to train BING\footnote{In fact, BING can generalize to generic object proposals without training as shown in \cite{hosang2014good}. Here we follow the original BING implementation.} with its default parameters, and test all the proposal algorithms on the test dataset. Microsoft COCO consists of 80 object categories with 82,081 images for training and 40,137 images for validation, leading to more than 2M annotated instances in total with ground-truth bounding boxes. Besides the amount of images and instances, the contents in images are more complex and challenging than those in VOC2007. On COCO we test all the proposal algorithms on the validation dataset using the same parameters as VOC2007 {\em without any retraining}.

We utilize the common intersection-over-union (IoU) overlap scoring function to measure the affinity of two bounding boxes, defined by the intersection area of two bounding boxes divided by their union. We measure our performance mainly in terms of (1) object detection recall (DR), (2) average best overlap (ABO) and mean average best overlap (MABO), and (3) computational time. We follow the PASCAL VOC challenge and use IoU overlap threshold $\eta=0.5$ by default for correct detection. 

We compare our method with \cite{Alexe2012pami}\footnote{\url{http://groups.inf.ed.ac.uk/calvin/objectness/}. We downloaded the proposals using either NMS or multinomial sampling. In our experiments we observed that for top 1,000 proposals NMS works better than multinomial sampling. Therefore we only reported the performance using NMS.}, \cite{Arbelaez_CVPR14}\footnote{\url{https://github.com/batra-mlp-lab/object-proposals}. We run the code for \cite{Alexe2012pami,Arbelaez_CVPR14,Blaschko2013b,Endres2012,Manen2013iccv,krahenbuhl2014geodesic,kk-lpo-15,Rantalankila14,Uijlings13,zitnickedge}.}, \cite{Blaschko2013b,Endres2012,Manen2013iccv,krahenbuhl2014geodesic,kk-lpo-15}, \cite{Rahtu_iccv11}\footnote{\url{http://www.cse.oulu.fi/CMV/Downloads/ObjectDetection/}}, \cite{Rantalankila14,Uijlings13,zitnickedge}, \cite{zhang2011proposal}\footnote{\url{https://zimingzhang.wordpress.com/source-code/}}, \cite{BingObj2014}\footnote{\url{https://github.com/varun-nagaraja/BING-Objectness}},  \cite{wang2015improving}\footnote{\url{http://3dimage.ee.tsinghua.edu.cn/cxz/mtse}}, and \cite{ren2015faster}\footnote{\url{https://github.com/ShaoqingRen/faster_rcnn}}. To evaluate the DR and MABO on VOC2007 test dataset, we download the precomputed proposals for \cite{Rahtu_iccv11}, \cite{Blaschko2013b}, \cite{Alexe2012pami} and \cite{wang2015improving} from the corresponding authors' websites. We use the default parameter setting for each method since they have been optimized for VOC2007, in general, expect for \cite{krahenbuhl2014geodesic} where we utilize the parameters $(180,9)$ as highlighted at the author's website. We utilize the evaluation code in \cite{zhang2011proposal, zhang2016object} for comparison. Specifically for each method we sort all the proposals based on their predicted scores in a descending order (or preserve their output orders if they do not have predicted scores) and keep at most top 1,000 proposals for computing DR and MABO. For RPN, we utilize GTX TITAN X for computation, and manage to tune the parameter in NMS to output around 1,000 proposals for comparison.

\begin{table}[t]\centering\footnotesize
\caption{\footnotesize{Performance comparison (\%) among different BING's derivatives.}}\label{tab:variants} \vspace{-3mm}
\setlength\tabcolsep{1.5pt}
\begin{tabular}{|c|cccc|cccc|c|c|}
\hline
\multirow{2}{*}{Methods} & \multicolumn{4}{c|}{DR, \# Prop., $\eta=0.5$} & \multicolumn{4}{c|}{DR, \# Prop., $\eta=0.7$} & MABO & Time\\
& 1 & 10 & 100 & 1000 & 1 & 10 & 100 & 1000 & (1000) & (ms)\\
\hline\hline
\multicolumn{11}{|c|}{VOC2007} \\
E-BING & \underline{\bf 20.2} & 34.1 & 65.8 & 91.3 & \underline{\bf 10.8} & 19.9 & 42.8 & 63.7 & 72.5 & \underline{\bf 1.7} \\
S-BING & 17.0 & 38.3 & 74.7 & \underline{\bf 95.1} & 7.9 & 17.9 & 49.0 & 76.4 & 76.8 & 2.6 \\
BING++ & 17.4 & \underline{\bf 42.1} & \underline{\bf 75.7} & 93.7 & 8.1 & \underline{\bf 20.4} & \underline{\bf 51.3} & \underline{\bf 77.3} & \underline{\bf 77.5} & 2.9 \\
\hline
\multicolumn{11}{|c|}{MS COCO} \\
E-BING & \underline{\bf 5.5} & 10.3 & 27.8 & 56.1 & \underline{\bf 2.7} & 5.6 & 15.2 & 31.5 & 50.1 & \underline{\bf 2.1} \\
S-BING & 4.8 & 13.8 & \underline{\bf 37.5} & \underline{\bf 64.8} & 2.3 & 5.8 & 20.6 & \underline{\bf 44.2} & \underline{\bf 56.9} & 3.0 \\
BING++ & 4.8 & \underline{\bf 14.8} & 37.2 & 62.6 & 2.3 & \underline{\bf 6.4} & \underline{\bf 21.2} & 43.1 & 56.0 & 3.7 \\
\hline
\end{tabular}
\end{table}

\begin{table}[t]\centering\footnotesize
	\caption{\footnotesize{Effect of image resize operation on performance (\%) in BING++.}}\label{tab:resize} \vspace{-3mm}
	\setlength\tabcolsep{1.5pt}
	\begin{tabular}{|c|cccc|cccc|c|c|}
		\hline
		\multirow{2}{*}{Methods} & \multicolumn{4}{c|}{DR, \# Prop., $\eta=0.5$} & \multicolumn{4}{c|}{DR, \# Prop., $\eta=0.7$} & MABO & Time\\
		& 1 & 10 & 100 & 1000 & 1 & 10 & 100 & 1000 & (1000) & (ms)\\
		\hline\hline
		\multicolumn{11}{|c|}{VOC2007} \\
		No+No & \underline{\bf 17.5} & 41.6 & 76.9 & \underline{\bf 95.0} & \underline{\bf 8.3} & 19.2 & 50.8 & \underline{\bf 78.8} & 78.0 & 4.2 \\
		Yes+No & 17.3 & 41.3 & 76.4 & 94.1 & 8.1 & 17.9 & 49.6 & 77.3 & 77.3 & 5.8 \\
		No+Yes & 17.3 & 41.9 & \underline{\bf 77.6} & 94.5 & 8.2 & 19.3 & \underline{\bf 51.4} & 78.7 & \underline{\bf 78.1} & 3.6 \\
		BING++ & 17.4 & \underline{\bf 42.1} & 75.7 & 93.7 & 8.1 & \underline{\bf 20.4} & 51.3 & 77.3 & 77.5 & \underline{\bf 2.9} \\
		\hline
		\multicolumn{11}{|c|}{MS COCO} \\
		No+No & \underline{\bf 5.1} & 14.5 & \underline{\bf 38.4} & \underline{\bf 65.5} & \underline{\bf 2.4} & \underline{\bf 6.6} & 21.2 & \underline{\bf 46.6} & \underline{\bf 57.9} & 11.2 \\
		Yes+No & 5.0 & 14.2 & 38.1 & 64.5 & \underline{\bf 2.4} & 5.8 & 20.4 & 44.8 & 57.0 & 12.1 \\
		No+Yes & 4.9 & \underline{\bf 14.8} & 38.2 & 63.9 & \underline{\bf 2.4} & 6.5 & \underline{\bf 22.0} & 45.4 & 57.1 & 7.2 \\
		BING++ & 4.8 & \underline{\bf 14.8} & 37.2 & 62.6 & 2.3 & 6.4 & 21.2 & 43.1 & 56.0 & \underline{\bf 3.7} \\
		\hline
	\end{tabular}
	\vspace{-3mm}
\end{table}

\subsection{Comparison on Derivatives of BING}\label{ssec:derivatives}

We refer to \textbf{\em E-BING} as edge based refinement (see Section~\ref{ssec:edge}), \textbf{\em S-BING} as segmentation based refinement (see Section~\ref{ssec:seg}), and \textbf{\em BING++} as sequential refinement using edge first and then segmentation (see Section~\ref{ssec:bing++}).

\begin{figure}[t]
	\centering
	\includegraphics[width = \linewidth]{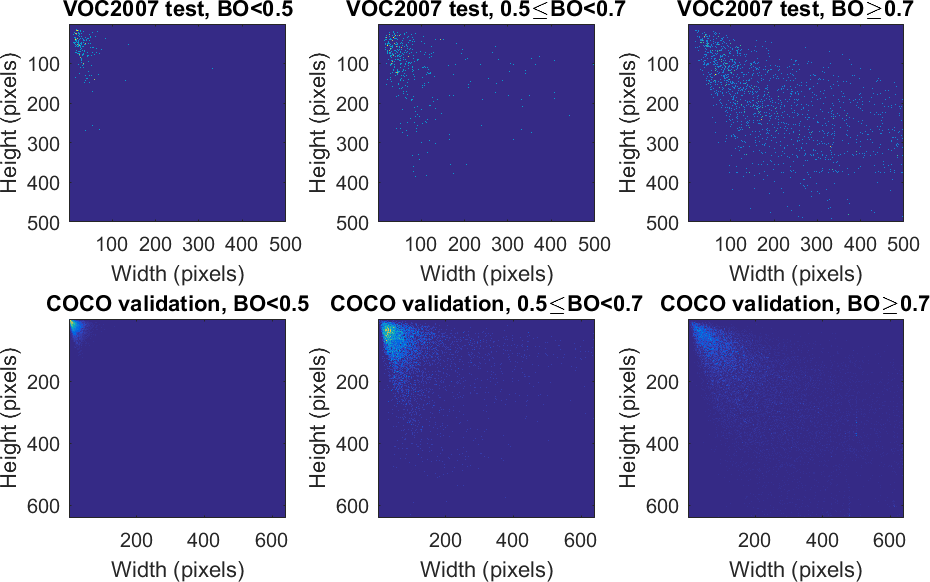}
	\vspace{-3mm}
	\caption{\footnotesize Distributions of objects based on their BO scores and the width and height of their ground-truth bounding boxes, given the proposals from BING++ as inputs. For larger objects BING++ works better, in general.}\label{fig:dist}
\end{figure}

\begin{figure}[t]
\centering
\includegraphics[width = \linewidth]{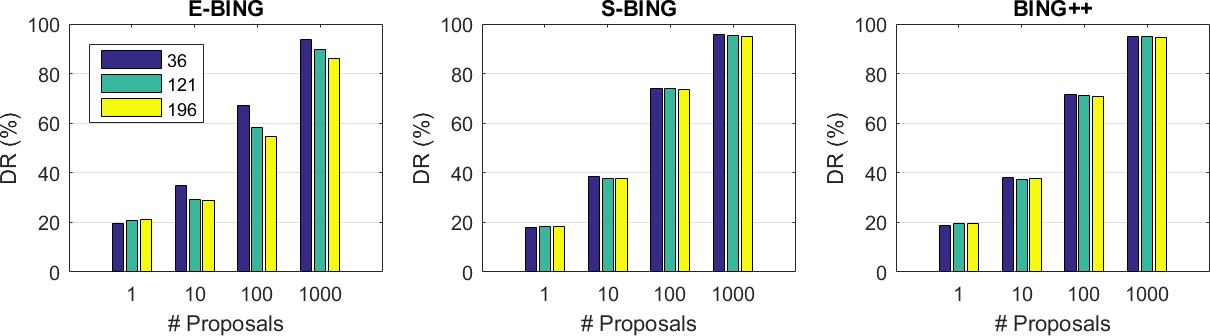}
\vspace{-3mm}
\caption{\footnotesize DR comparison on VOC2007 test dataset by varying the maximum number of quantized scales/aspect-ratios in BING.}\label{fig:param}
\vspace{-3mm}
\end{figure}

\begin{wrapfigure}{t}{0.45\linewidth} 
	\vskip-3ex
	\centering
	\includegraphics[width = \linewidth]{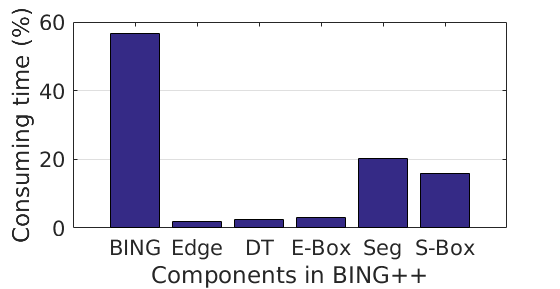}
	\caption{\footnotesize Timing distribution over components in BING++.}
	\label{fig:time}
\end{wrapfigure}

We first compare the performance of different BING's derivatives on VOC2007 test dataset and COCO validation dataset in Table \ref{tab:variants}, where the timing reported for all the methods is based on {\em multi-thread} computation on our server with two INTEL XEON E5 2696v2 CPU@2.50GHz. We observe that: (1) In terms of proposal quality, BING++ works the best, and S-BING is better than E-BING. (2) In terms of computation, E-BING is more efficient than S-BING and BING++. To see the contribution of each component in Alg.~\ref{alg:bing++} on the overall running time, we show the timing cost in percentage in Fig.~\ref{fig:time}. As we see, BING actually takes the largest portion of computation by 56.6\%, segmentation is ranked as the second by 20.1\%, and Alg. \ref{alg:seg} for segmentation based refinement is ranked as the third by 16.0\%. The extra computation for edge based refinement takes only 7.3\%. The increase of timing in Table \ref{tab:variants} roughly follows this distribution.

In BING++ image resize operation plays a very important role in reducing computational time. To show its effect on performance as well as running speed, we list all the comparison in Table \ref{tab:resize}, where ``Yes/No'' denotes with/without resize operation, the first ``Yes/No'' is for edge based refinement, and the second is for segmentation based refinement. In this context BING++ is equivalent to the ``Yes+Yes'' option. As we see here, all the four competitors perform similarly in terms of DR and MABO, especially when the number of proposals is small. This is because most of the objects with low best overlap (BO) scores are {\em small} in terms of number of pixels covered by the ground-truth bounding boxes, as shown in Fig. \ref{fig:dist}. Therefore, even ignoring small objects by resizing images has marginal effect on proposal quality for BING++, but results in significantly better computational efficiency.

\begin{table}[t]\centering\footnotesize
\caption{\footnotesize{DR (\%) and running time (s) comparison on VOC2007 test dataset.}} \label{tab:2} \vspace{-3mm}
\setlength\tabcolsep{1.8pt}
\begin{tabular}{|c|cccc|cccc|c|}
\hline
\multirow{2}{*}{Methods} & \multicolumn{4}{c|}{\# Prop., $\eta=0.5$} & \multicolumn{4}{c|}{\# Prop., $\eta=0.7$} & Time\\
& 1 & 10 & 100 & 1000 & 1 & 10 & 100 & 1000 & (s)\\
\hline\hline
Rahtu \cite{Rahtu_iccv11,Blaschko2013b} & 7.0 & 32.7 & 64.7 & 83.5 & 2.5 & 15.8 & 44.7 & 70.1 & 3.81\\
Objectness \cite{Alexe2012pami} & 17.3 & 49.5 & 75.8 & 92.0 & 7.4 & 23.4 & 37.6 & 43.1 & 3.83\\
CSVM \cite{zhang2011proposal} & 17.4 & 33.5 & 65.1 & 91.2 & 5.4 & 14.8 & 20.8 & 27.1 & 0.47\\
Sel. Search \cite{Uijlings13} & 9.7 & 37.3 & 71.5 & 93.5 & 4.1 & 19.7 & 49.0 & 80.0 & 10.64\\
Rand. Prim \cite{Manen2013iccv} & 8.6 & 35.0 & 70.4 & 90.3 & 3.5 & 17.3 & 45.1 & 73.4 & 0.79\\
Endres \cite{Endres2012} & \underline{{\bf 20.9}} & \underline{{\bf 55.2}} & \underline{{\bf 82.8}} & 90.1 & \underline{\bf 11.5} & \underline{\bf 35.0} & 58.0 & 73.0 & 11.67\\
Rantalankila \cite{Rantalankila14} & 0.1 & 0.9 & 16.2 & 85.6 & 0.0 & 0.4 & 8.5 & 67.5 & 23.72\\
GoP \cite{krahenbuhl2014geodesic} & 2.4 & 13.8 & 60.2 & 94.2 & 1.3 & 7.7 & 35.1 & 77.8 & 1.26\\
EdgeBox \cite{zitnickedge} & 17.8 & 45.8 & 75.4 & 95.1 & 9.5 & 30.9 & \underline{\bf 60.8} & \underline{\bf 85.1} & 0.25\\        
BING \cite{BingObj2014} & 18.2 & 37.3 & 73.0 & 95.2 & 7.3 & 16.9 & 24.5 & 29.1 & \underline{\bf 0.01}\\ 
MCG \cite{Arbelaez_CVPR14} & 18.5 & 44.2 & 65.7 & 86.5 & 9.4 & 26.9 & 49.1 & 70.1 & 18.97\\
LPO \cite{kk-lpo-15} & 18.5 & 38.0 & 75.5 & 94.4 & 8.0 & 18.0 & 49.2 & 76.8 & 1.43 \\
MTSE-BING \cite{wang2015improving} & 14.5 & 37.7 & 75.2 & \underline{\bf 95.3} & 7.0 & 18.1 & 47.2 & 78.1 & 0.15 \\
RPN \cite{ren2015faster} & 27.1 & 50.3 & 74.8 & 95.2 & 8.0 & 27.8 & 54.9 & 82.4 & 0.14 \\
\hline\hline
Rand. Uniform++ & 17.3 & 32.3 & 34.3 & 37.6 & 8.0 & 20.9 & 23.3 & 27.0 & \underline{\bf 0.01} \\
Rand. Gaussian++ & 17.2 & 34.8 & 38.2 & 43.0 & 8.0 & 22.8 & 27.0 & 32.1 & \underline{\bf 0.01} \\
Sliding Window++ & 17.2 & 37.4 & 47.0 & 50.8 & 8.0 & 23.6 & 34.6 & 38.4 & \underline{\bf 0.01} \\
\hline\hline
{\bf BING++} & 17.4 & 42.1 & 75.7 & 93.7 & 8.1 & 20.4 & 51.3 & 77.3 & 0.02 \\
\hline
\end{tabular}
\vspace{-3mm}
\end{table}

\begin{figure*}[t]
\begin{minipage}[b]{0.245\linewidth}
 \begin{center}
 \centerline{\includegraphics[width=1.05\columnwidth]{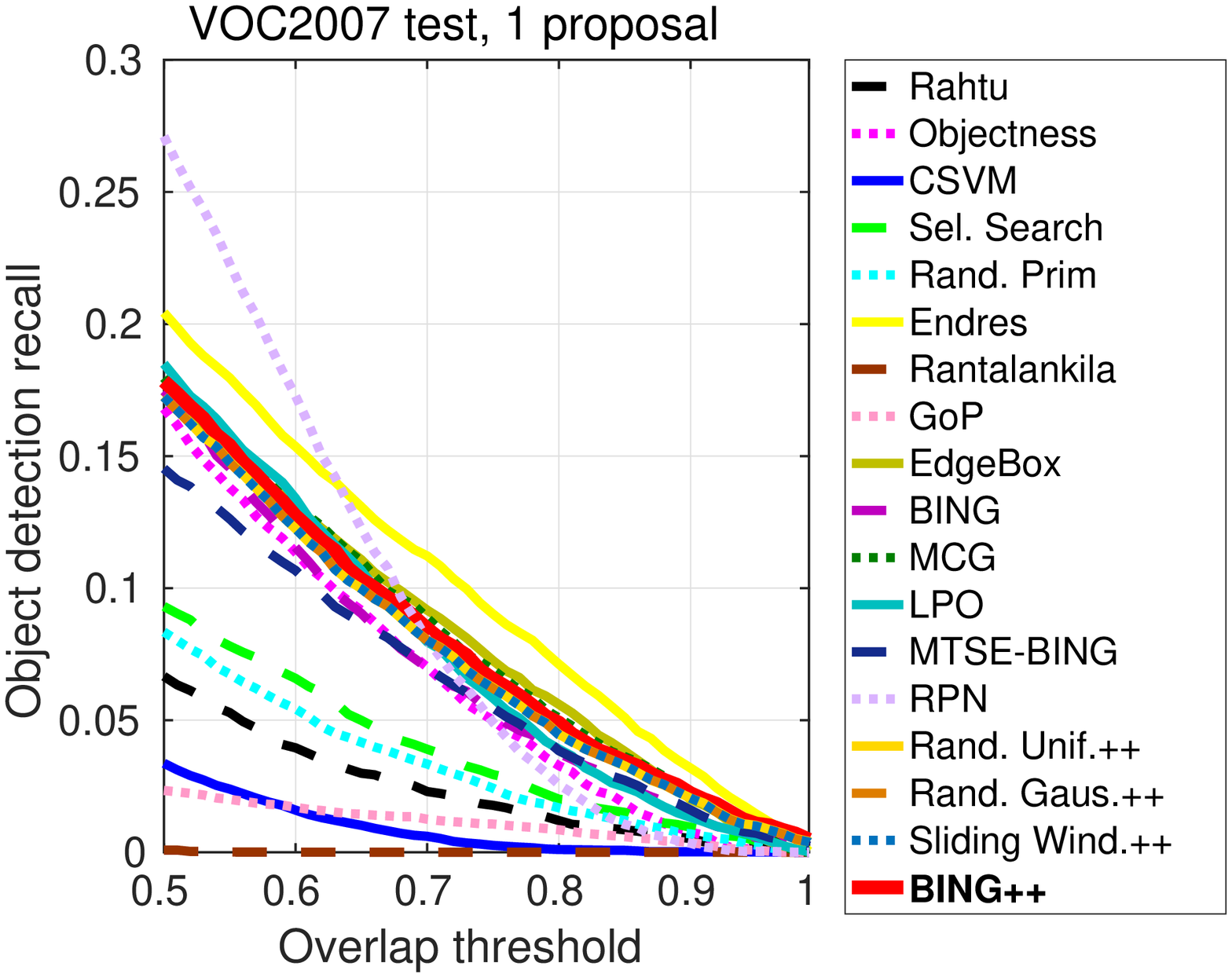}}
 \end{center}
\end{minipage}
\begin{minipage}[b]{0.245\linewidth}
 \begin{center}
 \centerline{\includegraphics[width=1.05\columnwidth]{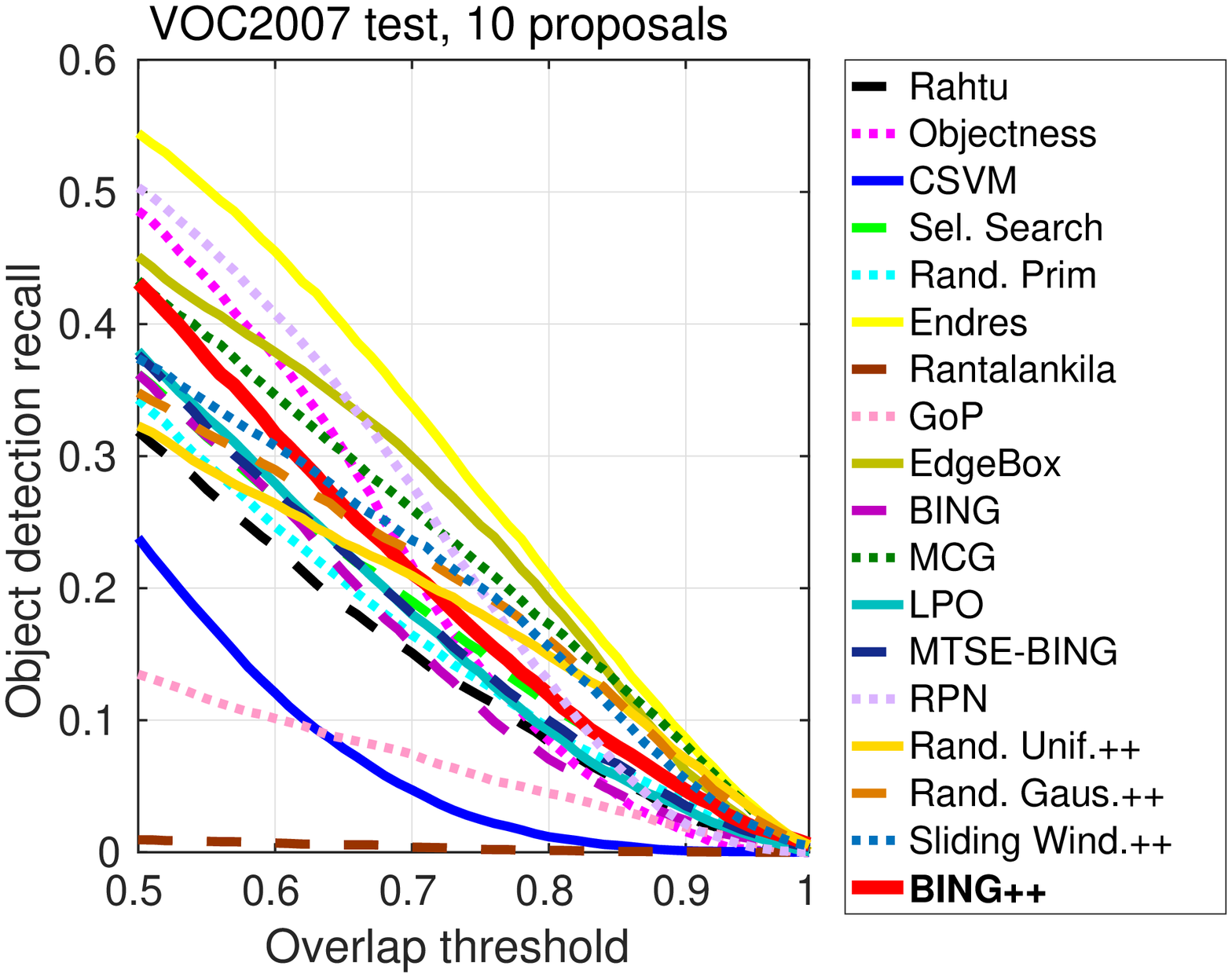}}
 \end{center}
\end{minipage}
\begin{minipage}[b]{0.245\linewidth}
 \begin{center}
 \centerline{\includegraphics[width=1.05\columnwidth]{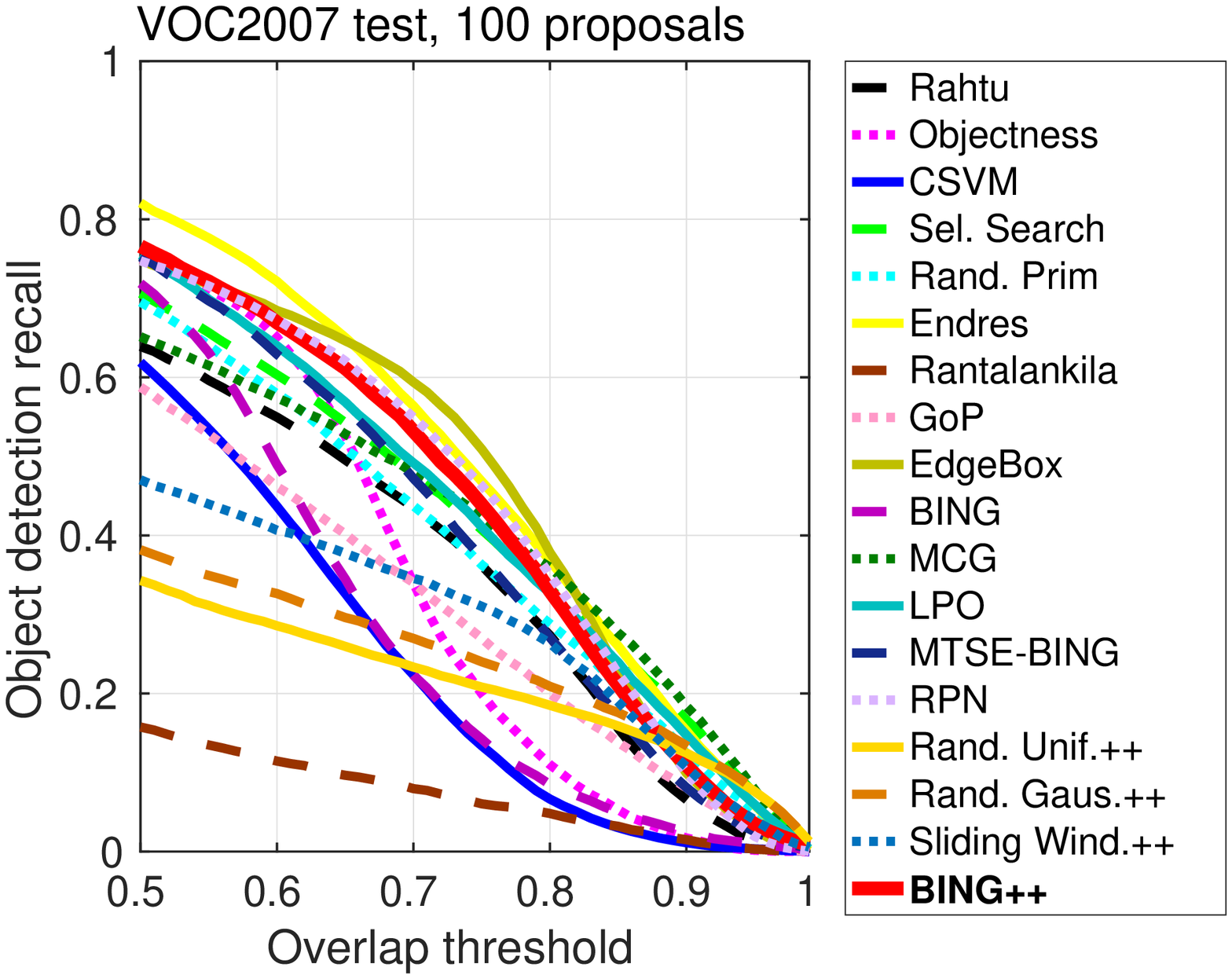}}
 \end{center}
\end{minipage}
\begin{minipage}[b]{0.245\linewidth}
 \begin{center}
 \centerline{\includegraphics[width=1.05\columnwidth]{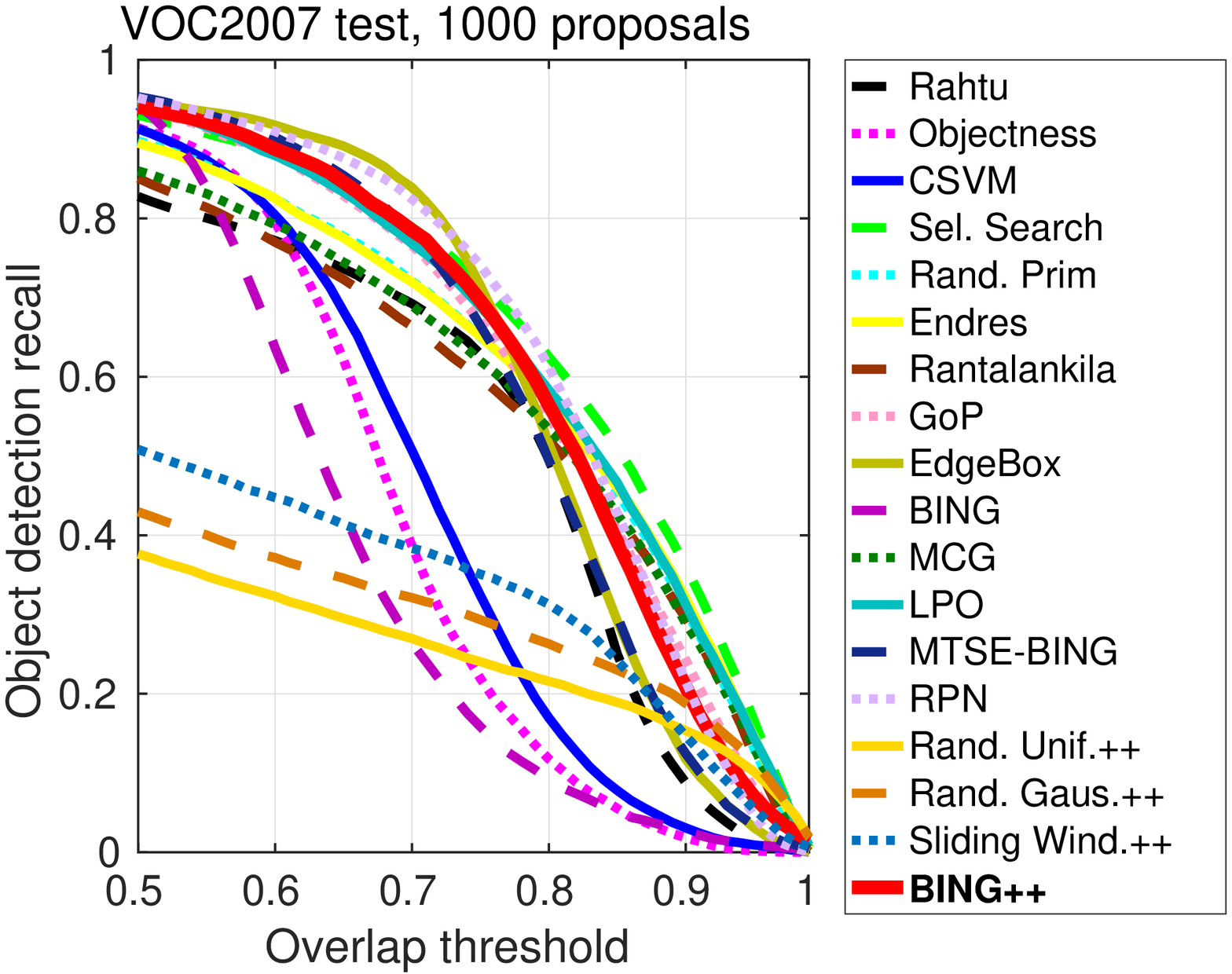}}
 \end{center}
\end{minipage}
\vspace{-9mm}
\caption{\footnotesize{Comparison of recall-overlap curves using different methods and numbers of proposals on VOC2007 test set.}}\label{fig:voc-recall-overlap}
\end{figure*}

\begin{table*}[t]\centering\footnotesize
\caption{\footnotesize{ABO \& MABO comparison (\%) between different proposal algorithms on VOC2007 test dataset using 1,000 proposals.}} \label{tab:ABO}\vspace{-3mm}
\setlength\tabcolsep{2.2pt}
\begin{tabular}{|c|cccccccccccccccccccc|c|}
\hline
 Methods & aer. & bic. & bird & boat & bot. & bus & car & cat & cha. & cow & din. & dog & hor. & mot. & per. & pot. & she. & sofa & tra. & tv & MABO \\ 
 \hline\hline
Rahtu \cite{Blaschko2013b,Rahtu_iccv11} & 72.6 & 74.5 & 69.0 & 69.7 & 49.0 & 79.9 & 67.2 & 82.6 & 62.9 & 72.9 & 80.5 & 80.6 & 78.5 & 74.5 & 65.7 & 58.9 & 69.8 & 82.3 & 79.5 & 73.8 & 72.2 \\
Objectness \cite{Alexe2012pami} & 67.3 & 69.0 & 65.6 & 64.4 & 57.1 & 72.3 & 65.3 & 72.8 & 64.8 & 67.6 & 73.3 & 71.1 & 70.9 & 67.6 & 63.5 & 61.4 & 65.6 & 74.7 & 70.6 & 65.5 & 67.5 \\
CSVM \cite{zhang2011proposal} & 70.4 & 70.2 & 67.6 & 66.7 & 58.9 & 73.6 & 67.6 & 76.5 & 64.9 & 68.0 & 74.1 & 74.7 & 72.7 & 71.4 & 66.6 & 62.3 & 67.9 & 75.6 & 74.5 & 66.7 & 69.6 \\
Sel. Search \cite{Uijlings13} & \underline{{\bf 83.5}} & \underline{{\bf 83.1}} & \underline{{\bf 80.1}} & \underline{{\bf 78.1}} & 62.8 & 85.2 & 77.7 & \underline{{\bf 90.7}} & \underline{{\bf 77.0}} & 83.2 & \underline{{\bf 88.6}} & \underline{{\bf 89.4}} & 82.9 & 81.9 & 72.6 & 71.2 & 80.4 & \underline{{\bf 90.3}} & \underline{{\bf 85.7}} & \underline{{\bf 83.7}} & \underline{{\bf 81.4}} \\
Rand. Prim \cite{Manen2013iccv} & 80.9 & 80.9 & 74.5 & 74.3 & 59.2 & 83.7 & 76.5 & 87.0 & 74.6 & 79.8 & 87.6 & 85.1 & 79.8 & 81.0 & 70.6 & 67.1 & 72.8 & 89.4 & 82.8 & 80.0 & 78.4 \\
Endres \cite{Endres2012} & 71.0 & 81.0 & 72.3 & 65.5 & 60.9 & 85.1 & 79.1 & 87.9 & 72.4 & 80.3 & 87.0 & 87.1 & 82.2 & \underline{{\bf 82.9}} & 70.5 & 67.4 & 76.2 & 89.7 & 84.5 & 79.1 & 78.1 \\
Rantalankila \cite{Rantalankila14} & 73.5 & 74.6 & 73.5 & 66.7 & 54.0 & 81.3 & 72.7 & 89.2 & 68.6 & 76.4 & 83.2 & 87.4 & 81.0 & 76.0 & 66.2 & 62.8 & 72.1 & 87.1 & 82.0 & 77.5 & 75.3 \\
GoP \cite{krahenbuhl2014geodesic} & 73.8 & 80.1 & 76.0 & 72.2 & 63.0 & \underline{{\bf 86.0}} & 80.3 & 88.2 & 75.9 & 81.3 & 85.8 & 85.7 & 79.9 & 79.1 & 73.5 & 71.1 & 78.7 & 88.4 & 82.3 & 82.3 & 79.2 \\
EdgeBoxes \cite{zitnickedge} & 76.8 & 81.6 & 78.5 & 76.7 & 65.8 & 83.9 & 76.8 & 82.3 & 76.4 & 82.2 & 80.9 & 83.6 & 81.3 & 80.9 & 73.6 & \underline{{\bf 71.7}} & \underline{{\bf 80.8}} & 82.5 & 79.8 & 81.4 & 78.9 \\
BING \cite{BingObj2014} & 65.5 & 66.0 & 64.0 & 62.3 & 60.6 & 66.5 & 64.4 & 69.9 & 62.6 & 65.1 & 69.5 & 68.3 & 65.9 & 65.7 & 63.8 & 62.4 & 64.6 & 69.0 & 68.6 & 63.4 & 65.4 \\
MCG \cite{Arbelaez_CVPR14} & 75.2 & 77.3 & 73.3 & 68.9 & 55.3 & 81.4 & 70.8 & 87.5 & 69.6 & 80.5 & 82.8 & 86.0 & 78.8 & 75.6 & 67.9 & 61.3 & 78.7 & 88.7 & 81.2 & 76.2 & 75.9 \\
LPO \cite{kk-lpo-15} & 74.9 & 79.9 & 76.9 & 72.9 & 61.4 & 86.4 & 80.4 & 89.1 & 74.5 & 82.0 & 85.1 & 86.9 & 82.4 & 81.7 & 73.0 & 71.5 & 79.4 & 88.7 & 85.3 & 81.6 & 79.7 \\
MTSE-BING \cite{wang2015improving} & 78.7 & 77.6 & 75.5 & 75.3 & 63.3 & 80.6 & 75.3 & 83.2 & 75.8 & 78.5 & 82.7 & 81.9 & 77.3 & 78.1 & 72.1 & 71.1 & 76.9 & 84.0 & 77.7 & 79.9 & 77.3 \\
RPN \cite{ren2015faster} & 76.8 & 81.9 & 77.6 & 76.4 & \underline{\bf 69.1} & 79.5 & \underline{\bf 81.8} & 84.4 & 76.0 & \underline{\bf 84.2} & 81.5 & 83.9 & \underline{\bf 83.5} & 80.7 & \underline{\bf 80.6} & 69.8 & 80.3 & 83.5 & 82.0 & 77.2 & 79.5 \\
 \hline\hline  
Rand. Uniform++ & 56.7 & 42.2 & 35.5 & 37.0 & 11.1 & 51.5 & 35.8 & 64.5 & 22.6 & 30.8 & 57.3 & 59.0 & 54.1 & 46.5 & 29.6 & 21.1 & 25.1 & 59.2 & 58.3 & 24.2 & 41.1 \\
Rand. Gaussian++ & 59.2 & 47.2 & 39.1 & 41.2 & 13.1 & 56.5 & 39.1 & 68.9 & 26.4 & 34.8 & 63.4 & 64.3 & 60.4 & 51.3 & 33.8 & 24.3 & 29.3 & 64.6 & 62.7 & 28.1 & 45.4 \\
Sliding Window++ & 62.0 & 54.6 & 45.5 & 45.2 & 16.1 & 62.2 & 44.5 & 73.7 & 33.7 & 40.9 & 71.1 & 69.3 & 62.6 & 56.2 & 36.4 & 29.0 & 37.1 & 73.0 & 67.0 & 37.1 & 50.9 \\
\hline\hline
{\bf BING++} & 79.5 & 78.5 & 76.6 & 75.2 & 60.0 & 81.5 & 75.5 & 85.3 & 72.4 & 78.2 & 83.6 & 84.0 & 79.7 & 79.2 & 70.7 & 68.7 & 77.9 & 85.5 & 79.8 & 77.2 & 77.5 \\
\hline
\end{tabular}
\vspace{-2mm}
\end{table*}

We also test the robustness of BING++ \wrt the quality of BING proposals by varying the maximum number of quantized scales/aspect-ratios (\ie 36, 121, 196) in BING to generate different proposals (see the details in \cite{zhang2011proposal}). We then feed all these BING proposals into the three derivatives. We show the DR comparison in Fig. \ref{fig:param} (for MABO we observe similar behavior for each method). As we see E-BING (as well as BING) is actually sensitive to the parameter, but S-BING and BING++ are not. This is because segments are much stronger clues to estimate object boundaries than edges, leading to robust performance.

%

\begin{table}[t]\centering\scriptsize
\caption{\footnotesize{DR (\%), MABO (\%) and running time (s) comparison on VOC2007 test dataset with images resized to $1/9$ of their original sizes.}} \label{tab:edge_resize} \vspace{-3mm}
\setlength\tabcolsep{1.8pt}
\begin{tabular}{|c|cccc|cccc|c|c|}
\hline
\multirow{2}{*}{Methods} & \multicolumn{4}{c|}{\# Prop., $\eta=0.5$} & \multicolumn{4}{c|}{\# Prop., $\eta=0.7$} & MABO & Time\\
& 1 & 10 & 100 & 1000 & 1 & 10 & 100 & 1000 & & (s)\\
\hline\hline
Rahtu \cite{Rahtu_iccv11,Blaschko2013b} & 17.5 & 34.2 & 59.6 & 79.6 & 8.2 & 21.7 & 44.8 & 68.2 & 69.8 & 1.06 \\
Sel. Search \cite{Uijlings13} & 10.4 & 37.4 & 71.6 & \underline{\bf 94.4} & 4.5 & 20.1 & 49.0 & \underline{\bf 77.7} & \underline{\bf 79.7} & 5.13 \\
Rand. Prim \cite{Manen2013iccv} & 10.0 & 39.6 & 69.5 & 71.9 & 3.9 & 20.7 & 48.2 & 53.2 & 65.1 & 0.11 \\
Endres \cite{Endres2012} & 21.3 & 50.0 & 66.1 & 66.7 & \underline{\bf 11.1} & \underline{\bf 29.7} & 45.8 & 46.9 & 60.3 & 2.31 \\
Rantalankila \cite{Rantalankila14} & 0.0 & 0.0 & 6.9 & 39.2 & 0.0 & 0.0 & 0.8 & 10.6 & 41.5 & 9.57 \\
GoP \cite{krahenbuhl2014geodesic} & 1.0 & 5.6 & 36.7 & 89.2 & 0.5 & 2.2 & 17.2 & 66.5 & 73.9 & 0.69 \\
EdgeBox \cite{zitnickedge} & 18.6 & 39.2 & 64.1 & 80.9 & 8.0 & 22.6 & 46.1 & 65.0 & 67.9 & 0.09 \\       
MCG \cite{Arbelaez_CVPR14} & 20.2 & \underline{\bf 51.9} & \underline{\bf 77.0} & 91.4 & 10.3 & 27.6 & \underline{\bf 53.0} & 74.3 & 78.8 & 4.55 \\
LPO \cite{kk-lpo-15} & 17.6 & 37.3 & 70.3 & 89.5 & 7.2 & 16.5 & 43.5 & 68.8 & 75.5 & 1.09 \\
RPN \cite{ren2015faster} & \underline{\bf 22.0} & 40.8 & 58.6 & 82.7 & 5.6 & 19.6 & 40.7 & 67.7 & 70.6 & 0.15 \\
\hline\hline
{\bf BING++} & 17.4 & 42.1 & 75.7 & 93.7 & 8.1 & 20.4 & 51.3 & 77.3 & 77.5 & \underline{\bf 0.02} \\
\hline
\end{tabular}
\end{table}

\begin{table}[t]\centering\scriptsize
\caption{\footnotesize{DR (\%), MABO (\%) and running time (s) comparison on VOC2007 test dataset with images resized to $360\times400$ pixels.}} \label{tab:seg_resize} \vspace{-3mm}
\setlength\tabcolsep{1.8pt}
\begin{tabular}{|c|cccc|cccc|c|c|}
\hline
\multirow{2}{*}{Methods} & \multicolumn{4}{c|}{\# Prop., $\eta=0.5$} & \multicolumn{4}{c|}{\# Prop., $\eta=0.7$} & MABO & Time\\
& 1 & 10 & 100 & 1000 & 1 & 10 & 100 & 1000 & & (s)\\
\hline\hline
Rahtu \cite{Rahtu_iccv11,Blaschko2013b} & 18.5 & 34.0 & 62.4 & 82.9 & 8.7 & 21.3 & 45.8 & 70.2 & 71.5 & 2.34 \\
Sel. Search \cite{Uijlings13} & 10.4 & 37.6 & 72.3 & 95.1 & 4.2 & 20.3 & 50.3 & 79.8 & 81.3 & 6.68 \\
Rand. Prim \cite{Manen2013iccv} & 9.0 & 35.0 & 71.0 & 89.6 & 3.8 & 17.8 & 46.8 & 72.6 & 77.9 & 0.56 \\
Endres \cite{Endres2012} & 21.6 & \underline{\bf 55.0} & 82.3 & 89.7 & \underline{\bf 11.9} & \underline{\bf 34.7} & 58.2 & 72.2 & 77.5 & 9.50 \\
Rantalankila \cite{Rantalankila14} & 0.0 & 0.1 & 2.5 & 56.8 & 0.0 & 0.0 & 0.3 & 25.7 & 52.84 & 12.0 \\
GoP \cite{krahenbuhl2014geodesic} & 2.3 & 14.8 & 61.7 & 93.9 & 1.2 & 7.6 & 36.4 & 76.5 & 78.6 & 1.32 \\
EdgeBox \cite{zitnickedge} & 19.7 & 41.1 & 66.9 & 90.0 & 8.6 & 24.7 & 50.5 & 74.9 & 73.5 & 0.16 \\       
MCG \cite{Arbelaez_CVPR14} & 17.9 & 52.1 & \underline{\bf 82.6} & \underline{\bf 95.3} & 9.8 & 29.3 & \underline{\bf 59.9} & \underline{\bf 81.9} & \underline{\bf 82.7} & 16.47 \\
LPO \cite{kk-lpo-15} & 18.4 & 38.3 & 74.7 & 88.5 & 8.1 & 18.3 & 48.2 & 67.7 & 75.4 & 2.26 \\
RPN \cite{ren2015faster} & \underline{\bf 28.6} & 50.2 & 74.5 & 94.9 & 10.2 & 29.8 & 54.7 & 81.8 & 79.7 & 0.13 \\
\hline\hline
{\bf BING++} & 17.4 & 42.1 & 75.7 & 93.7 & 8.1 & 20.4 & 51.3 & 77.3 & 77.5 & \underline{\bf 0.02} \\
\hline
\end{tabular}
\vspace{-3mm}
\end{table}

\subsection{Benchmark Comparison (I): VOC2007}
We first compare our BING++ with other proposal algorithms using DR \vs IoU overlap threshold in Fig. \ref{fig:voc-recall-overlap}. We also implement three other baseline methods by replacing BING proposals in BING++ with bounding boxes sampled by (1) Random Uniform, (2) Random Gaussian, or (3) Sliding Window\footnote{We used the published code at \url{https://github.com/hosang/detection-proposals} for \cite{Hosang2015Pami} to sample 1,000 proposals.}. These sampling methods have ignorable running time, leading to faster speed than BING++, as shown in Table \ref{tab:2}, yet much worse proposal quality. Overall, our BING++ behaves similarly to many other competitive proposal algorithms such as selective search, edgeBoxes, and GoP, especially when the number of proposals is sufficiently large (\eg 100 or 1,000). Note that when the proposals are sufficient, there are significant performance gaps between BING and our BING++, indicating that BING++ achieves huge improvement on DR over BING. 

To quantify these plots, we list the corresponding numbers as well as the average {\em single-thread} computational time of each method in Table \ref{tab:2}, which are called from MATLAB, except RPN. In both cases with IoU threshold equal to 0.5 or 0.7, BING++ can always achieve similar performance to the best ones. However, it is quite notable that BING++ is much faster than other competitive proposal algorithms. For instance, BING++ is 500 times faster than selective search. Also by comparing BING++ with BING 
using IoU threshold equal to 0.7 and 1000 proposals, there is huge improvement again from 29.1\% to 77.3\%. This well demonstrates the capability of BING++ for generating high quality object proposals. 

\begin{table*}[t]\centering\footnotesize
\caption{\footnotesize{AP \& mAP comparison (\%) for object detection using fast R-CNN on VOC2007 test dataset with 1,000 proposals with IoU threshold 0.5.}} \label{tab:det_full}\vspace{-3mm}
\setlength\tabcolsep{2.1pt}
\begin{tabular}{|c|cccccccccccccccccccc|c|}
\hline
Methods (\# Proposals) & aer. & bic. & bird & boat & bot. & bus & car & cat & cha. & cow & din. & dog & hor. & mot. & per. & pot. & she. & sofa & tra. & tv & mAP \\
\hline\hline
Objectness \cite{Alexe2012pami} (1,000) & 61.1& 64.9& 54.0& 38.1& 25.6& 71.8& 69.5& 75.7& 29.7& 66.0& 57.3& 68.2& 76.8& 61.9& 51.4& 19.4& 51.2& 53.9& 65.3& 59.9& 56.1\\
Sel. Search \cite{Uijlings13} (999) & 71.9& 78.3& 68.2& 53.6& 35.4& 78.9& 75.5& 85.6& 42.4& 75.7& 70.0& 80.5& 82.9& 72.5& 66.3& 35.0 & 68.4& 66.9& \underline{\bf 77.9}& 68.4& 67.7\\
Rand. Prim\cite{Manen2013iccv}  (852) & \underline{\bf 76.2}& 79.0& 63.0& 54.8 & 30.1& 80.9& 72.2& 84.6& 39.5& 73.5& 70.1& 77.2& 80.6& 72.4& 64.1& 28.3& 59.3& 69.5& 75.8,& 66.5& 65.9\\
Endres \cite{Endres2012} (830) & 66.9& 76.2& 67.2& 47.8& 33.4& 80.8& 75.0& 84.3& 36.5& 71.6& 68.0& 78.1& 77.9& 71.7& 63.7& 27.6& 63.7& \underline{\bf 69.8}& 76.2& 67.3& 65.2\\
Rantalankila \cite{Rantalankila14}  (856) & 68.1& 70.2& 64.3& 49.0& 27.2& 78.6& 69.8& 82.4& 33.1& 71.4& 67.0& 79.7& 78.3& 69.1& 58.9& 26.0& 61.9& 65.4& 76.9& 65.8& 63.2\\
GoP \cite{krahenbuhl2014geodesic} (992) & 66.8& 80.6& \underline{\bf 68.6}& 50.9& 36.4& 82.7& \underline{\bf 80.6} & 85.6 & 39.3 & 74.4& \underline{\bf 70.9}& \underline{\bf 82.2} & \underline{\bf 86.2} & \underline{\bf 78.9}& 68.2& 29.9& 66.4& 69.5& 74.7& 69.7 & 68.1 \\
EdgeBoxes \cite{zitnickedge}  (991) & 64.2& 79.2& 66.8& 53.6& 38.0 & \underline{\bf 83.8}& 78.4& 84.4& 40.1& \underline{\bf 78.3}& 65.5& 81.9& 83.3& 76.5& 69.6 & 34.3& \underline{\bf 69.5} & 65.1& 73.0& 67.6& 67.7 \\
BING \cite{BingObj2014} (1,000) & 56.7& 63.1& 55.3& 37.4& 34.8& 70.6& 69.9& 70.6& 27.7& 61.7& 45.6& 63.5& 72.5& 62.9& 57.5& 21.8& 52.0& 46.8& 64.9& 57.1& 54.6 \\
LPO \cite{kk-lpo-15}(823) & 66.7& \underline{\bf 81.9}& 68.4& 52.5& 33.5& 81.7& 78.4& \underline{\bf 85.9} & 40.5& 76.4& 67.6& 81.5& 85.7& 74.4& 68.1& 34.7& 67.6& 66.4& 74.6& 67.9& 67.7 \\
MCG (1,000) \cite{Arbelaez_CVPR14} & 67.0& 74.0& 64.9& 49.9& 32.5& 77.4& 64.9& 83.7& 34.9& 72.1& 69.9& 78.5& 78.0& 69.4& 60.5& 28.8& 64.2& 67.5& 76.4& 64.2& 63.9\\
MTSE-BING \cite{wang2015improving} (996) & 70.1& 80.6& 65.5& 54.0& 36.7& 81.7& 79.3& 84.6& 40.1& 78.2& 64.2& 79.5& 84.2& 76.8& 69.1& 33.8& 67.2& 67.5& 73.5& 67.9& 67.7 \\
RPN \cite{ren2015faster} & 68.6 & 78.9 & 67.4 & \underline{\bf 54.9} & \underline{\bf 52.4} & 75.2 & 79.7 & 79.1 & \underline{\bf 49.5} & 75.5 & 66.5 & 76.6 & 80.9 & 77.2 & \underline{\bf 77.4} & \underline{\bf 41.7} & 68.2 & 64.1 & 72.8 & \underline{\bf 72.4} & \underline{\bf 69.0} \\
\hline\hline  
No + No (1,000) & 69.8 & 75.9 & 67.5 & 53.7 & 38.2 & 78.6 & 76.3 & 82.6 & 41.1 & 74.6 & 65.7 & 79.0 & 83.3 & 75.9 & 66.8 & 35.3 & 65.1 & 66.0 & 72.1 & 65.0 & 66.6 \\
Yes + No (1,000) & 67.0 & 75.1 & 65.1 & 53.7 & 36.9 & 81.7 & 75.7 & 82.7 & 40.2 & 72.8 & 64.7 & 79.6 & 81.3 & 75.4 & 66.2 & 34.2 & 66.2 & 65.6 & 72.6 & 64.6 & 66.1 \\
No + Yes (1,000) & 68.9 & 75.4 & 63.9 & 53.3 & 36.2 & 77.9 & 76.9 & 83.0 & 38.5 & 74.9 & 65.2 & 75.3 & 81.9 & 74.8 & 65.9 & 34.5 & 68.5 & 65.8 & 69.0 & 66.1 & 65.8 \\
{\bf BING++} (1,000) & 66.3 & 75.4 & 65.1 & 52.7 & 35.8 & 79.5 & 75.8 & 80.5 & 40.8 & 72.5 & 66.3 & 75.7 & 81.5 & 73.7 & 65.6 & 35.0 & 63.4 & 63.4 & 74.7 & 64.2 & 65.4 \\
\hline
\end{tabular}
\end{table*}

\begin{figure}[t]
\begin{minipage}[b]{0.495\columnwidth}
 \begin{center}
 \centerline{\includegraphics[width=\columnwidth]{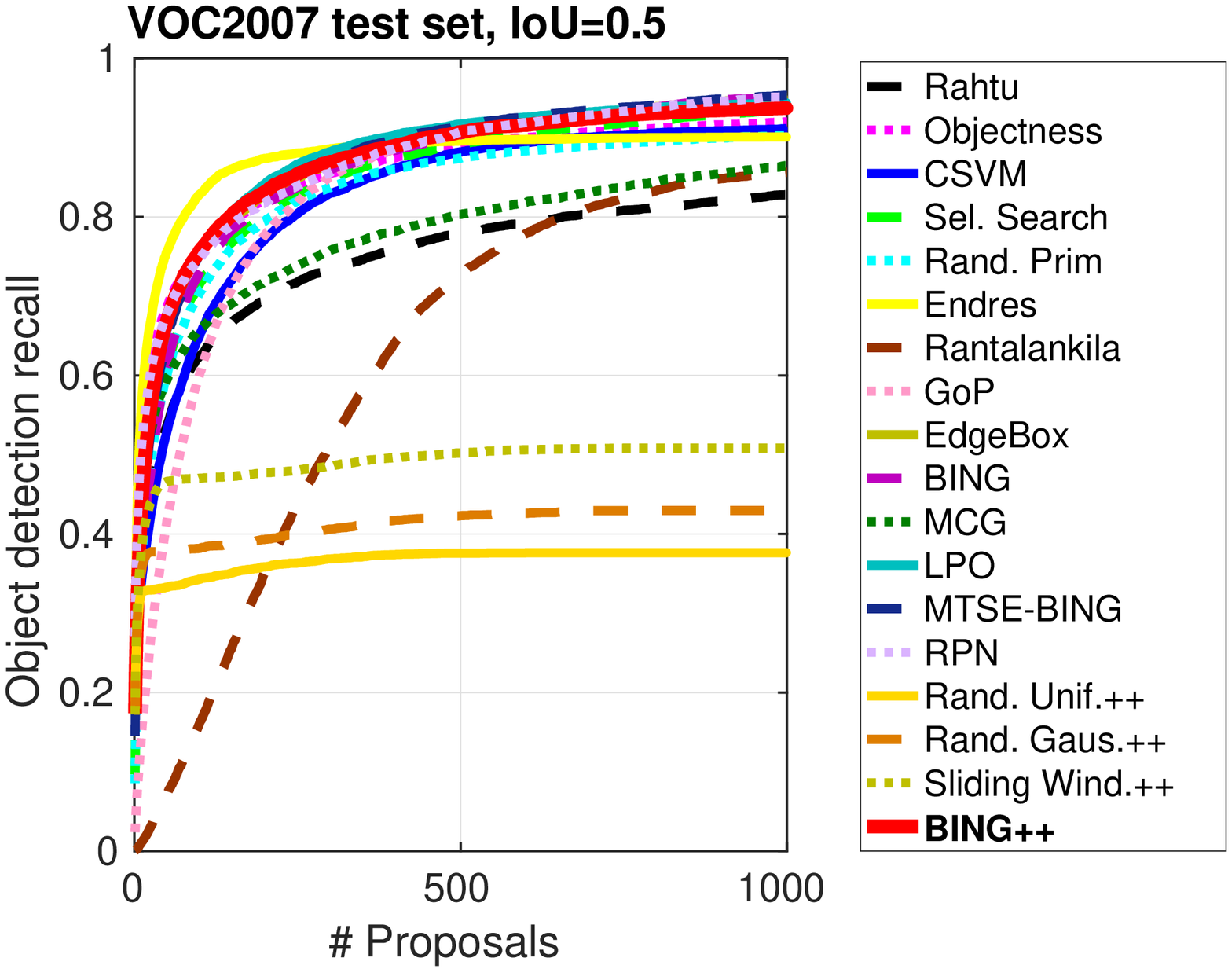}}
 \centerline{\footnotesize{(a)}}
 \end{center}
\end{minipage}
\begin{minipage}[b]{0.495\columnwidth}
 \begin{center}
 \centerline{\includegraphics[width=\columnwidth]{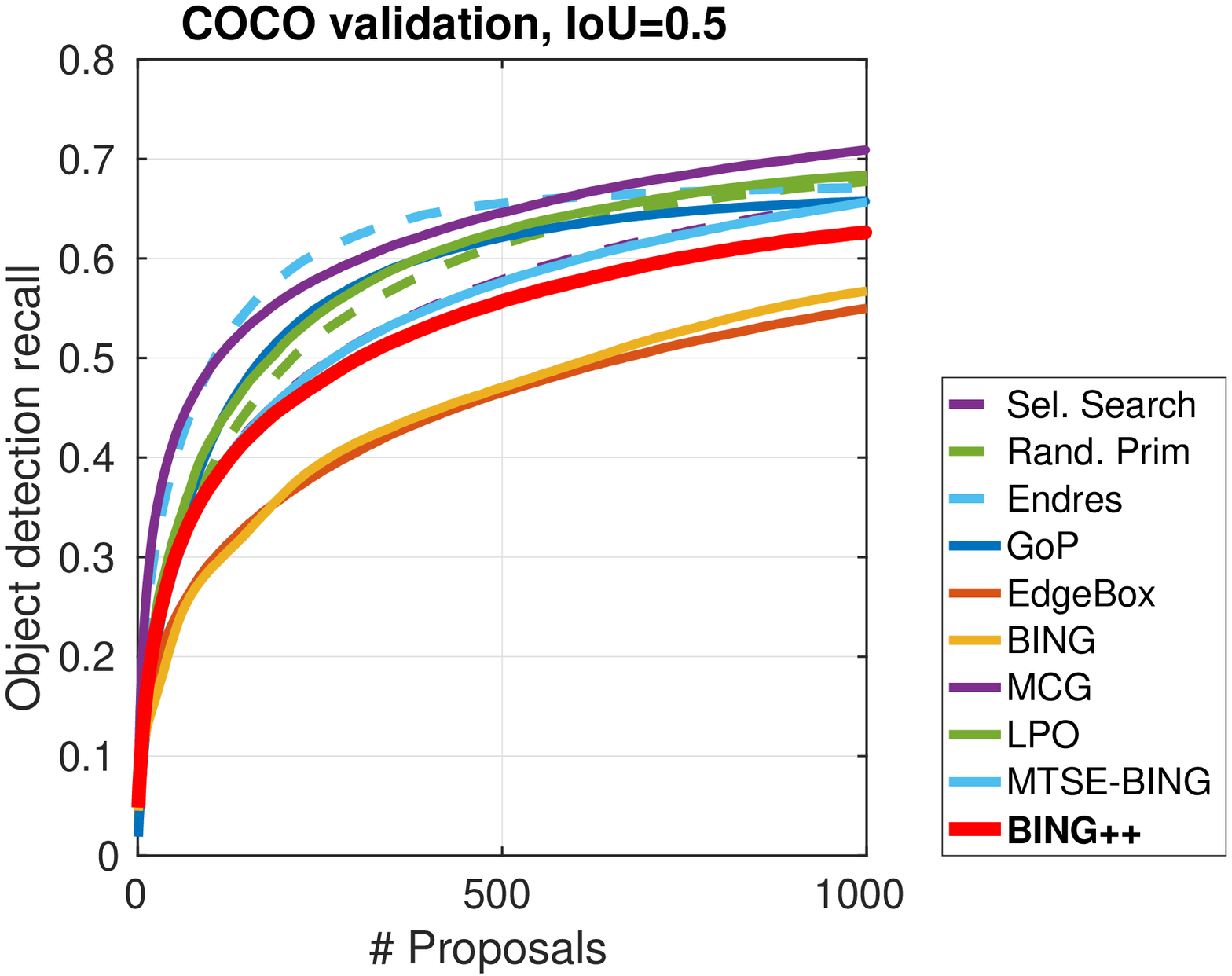}}
 \centerline{\footnotesize{(b)}} 
 \end{center}
\end{minipage}
\vspace{-9mm}
\caption{\footnotesize{Comparison of DR \vs number of proposals on VOC2007 test dataset {\bf (a)} and MS COCO validation dataset {\bf (b)}, respectively.}}\label{fig:dr}
\vspace{-3mm}
\end{figure}

\begin{figure*}[t]
\begin{minipage}[b]{0.162\linewidth}
 \begin{center}
 \centerline{\includegraphics[width=\columnwidth]{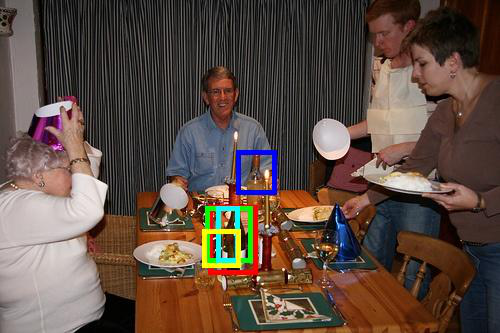}}
 \end{center}
\end{minipage}
\begin{minipage}[b]{0.162\linewidth}
 \begin{center}
 \centerline{\includegraphics[width=\columnwidth]{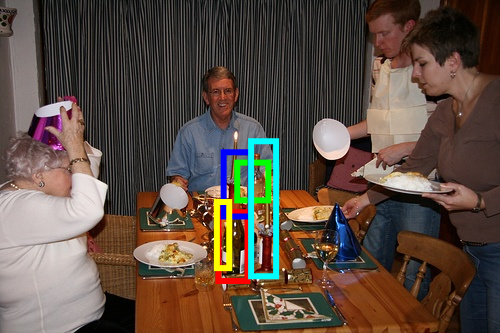}}
 \end{center}
\end{minipage}
\begin{minipage}[b]{0.162\linewidth}
 \begin{center}
 \centerline{\includegraphics[width=\columnwidth]{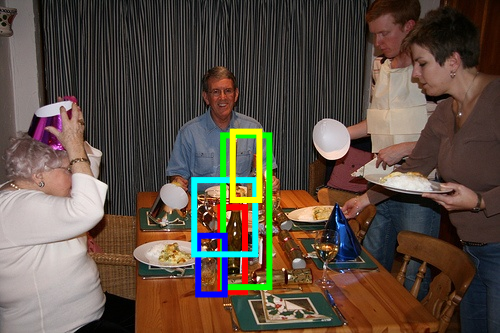}}
 \end{center}
\end{minipage}
\begin{minipage}[b]{0.162\linewidth}
 \begin{center}
 \centerline{\includegraphics[width=\columnwidth]{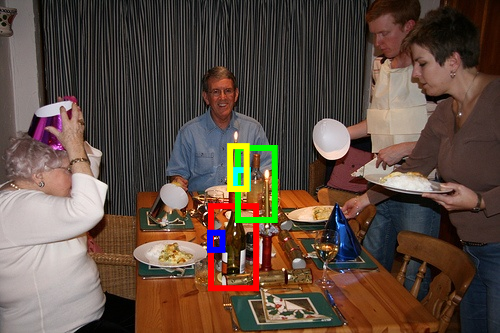}}
 \end{center}
\end{minipage}
\begin{minipage}[b]{0.162\linewidth}
 \begin{center}
 \centerline{\includegraphics[width=\columnwidth]{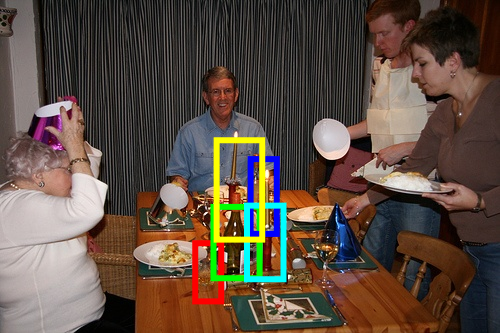}}
 \end{center}
\end{minipage}
\begin{minipage}[b]{0.162\linewidth}
 \begin{center}
 \centerline{\includegraphics[width=\columnwidth]{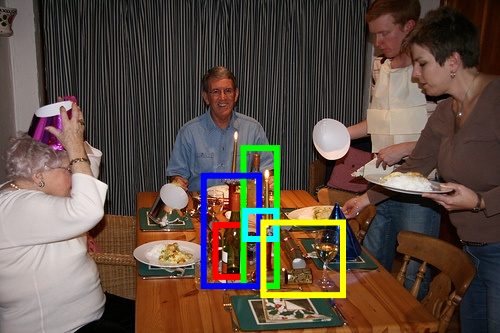}}
 \end{center}
\end{minipage}
\begin{minipage}[b]{0.162\linewidth}
 \begin{center}
 \centerline{\includegraphics[width=\columnwidth]{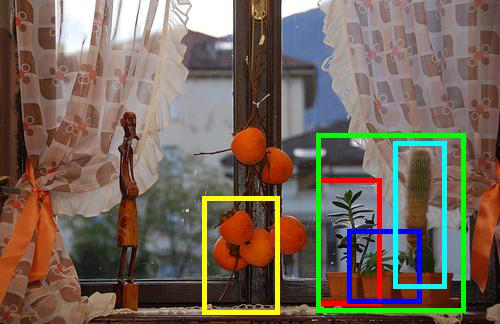}}
 \end{center}
\end{minipage}
\begin{minipage}[b]{0.162\linewidth}
 \begin{center}
 \centerline{\includegraphics[width=\columnwidth]{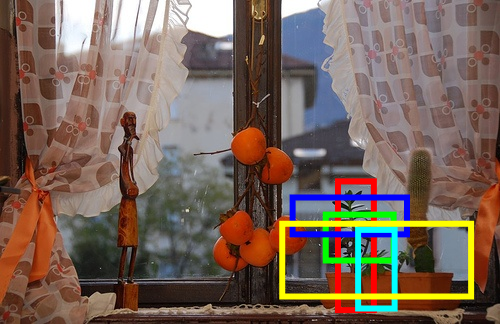}}
 \end{center}
\end{minipage}
\begin{minipage}[b]{0.162\linewidth}
 \begin{center}
 \centerline{\includegraphics[width=\columnwidth]{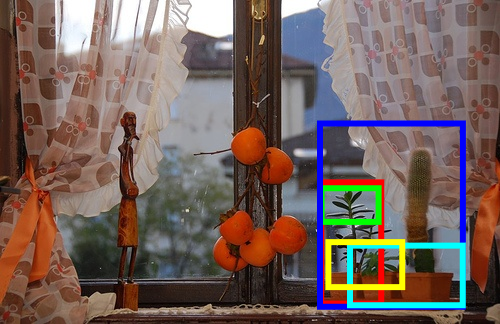}}
 \end{center}
\end{minipage}
\begin{minipage}[b]{0.162\linewidth}
 \begin{center}
 \centerline{\includegraphics[width=\columnwidth]{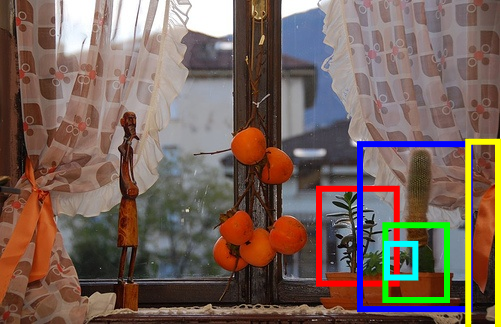}}
 \end{center}
\end{minipage}
\begin{minipage}[b]{0.162\linewidth}
 \begin{center}
 \centerline{\includegraphics[width=\columnwidth]{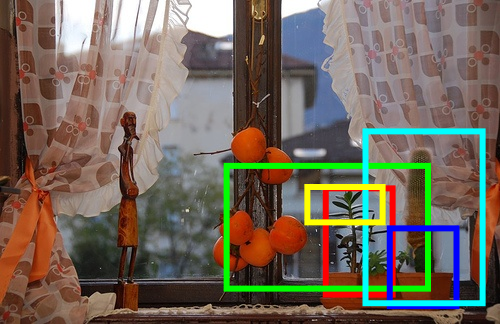}}
 \end{center}
\end{minipage}
\begin{minipage}[b]{0.162\linewidth}
 \begin{center}
 \centerline{\includegraphics[width=\columnwidth]{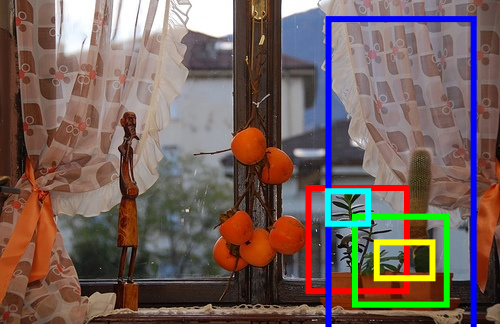}}
 \end{center}
\end{minipage}
\begin{minipage}[b]{0.162\linewidth}
 \begin{center}
 \centerline{\includegraphics[width=\columnwidth]{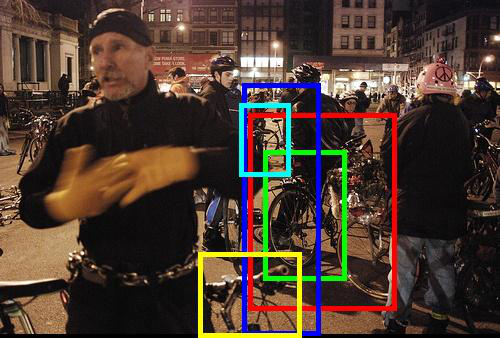}}
 \centerline{\footnotesize{{\bf (a) BING++}}}
 \end{center}
\end{minipage}
\begin{minipage}[b]{0.162\linewidth}
 \begin{center}
 \centerline{\includegraphics[width=\columnwidth]{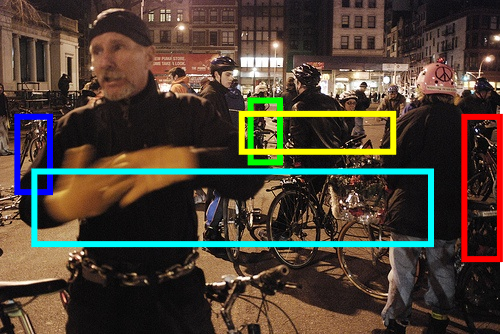}}
 \centerline{\footnotesize{(b) BING}}
 \end{center}
\end{minipage}
\begin{minipage}[b]{0.162\linewidth}
 \begin{center}
 \centerline{\includegraphics[width=\columnwidth]{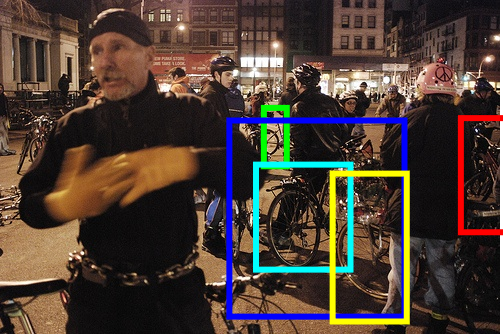}}
 \centerline{\footnotesize{(c) EdgeBoxes}}
 \end{center}
\end{minipage}
\begin{minipage}[b]{0.162\linewidth}
 \begin{center}
 \centerline{\includegraphics[width=\columnwidth]{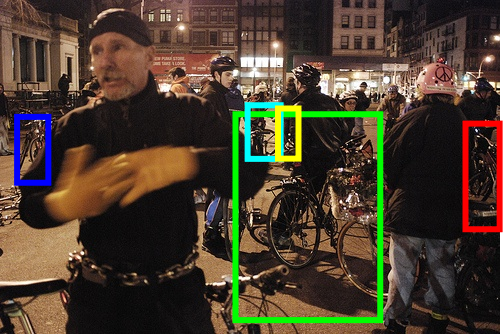}}
 \centerline{\footnotesize{(d) LPO}}
 \end{center}
\end{minipage}
\begin{minipage}[b]{0.162\linewidth}
 \begin{center}
 \centerline{\includegraphics[width=\columnwidth]{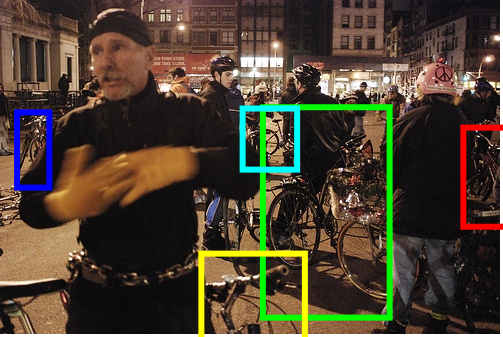}}
 \centerline{\footnotesize{(e) MTSE-BING}}
 \end{center}
\end{minipage}
\begin{minipage}[b]{0.162\linewidth}
 \begin{center}
 \centerline{\includegraphics[width=\columnwidth]{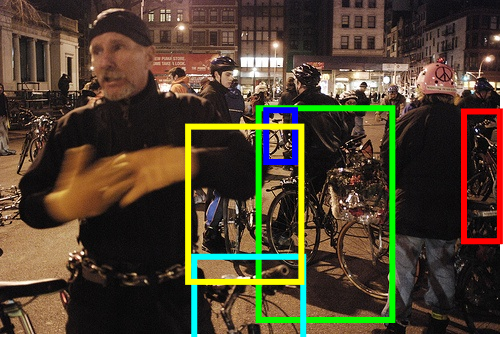}}
 \centerline{\footnotesize{(f) Selective Search}}
 \end{center}
\end{minipage}
\vspace{-9mm}
\caption{\footnotesize{Illustration of detection results on VOC2007 test dataset for some comparative algorithms using classes of (top) bottle, (middle) potted plants, and (bottom) bicycle. In each image, we show top-5 detections for the class, colored by red, green, blue, cyan, and yellow.}}\label{fig:samples}
\vspace{-3mm}
\end{figure*}


In order to understand the effect of image down-sampling on performance of different competitors, we resize the images to either $1/3\times 1/3=1/9$ of their original sizes (used in edge-based refinement in BING++) or the fixed size of $360\times400$ pixels (used in segmentation-based refinement in BING++), respectively, and repeat the same experiments in Table~\ref{tab:2} and Table~\ref{tab:ABO} using the {\em default} parameters for each algorithm without fine-tuning. The results are listed in Table \ref{tab:edge_resize} and Table~\ref{tab:seg_resize}, respectively. As we see: (1) Among the comparison our BING++ is still the fastest as well as achieving comparable performance with the state-of-the-art. (2) The running time of RPN is quite similar in various experiments, because images are rescaled to the same size as inputs to the neural network. (3) For the rest of competitors, generally speaking, the running time is improved, and the smaller the images are, the faster the methods can run. However, the DR and MABO become worse, in general. In Table \ref{tab:edge_resize} even if EdgeBox can run 4.5 times slower than BING++, its DRs are 12.8\% (with $\eta=0.5$ and 1,000 proposals) and 12.3\% (with $\eta=0.7$ and 1,000 proposals) lower than those of BING++, and its MABO is 9.6\% lower than that of BING++.

Next we show the comparison of DR \vs number of proposals in Fig.~\ref{fig:dr}(a). Clearly with $\eta=0.5$, BING++ performs among the top, which is consistent with Table~\ref{tab:2}. We also list our ABO and MABO comparison in Table~\ref{tab:ABO}. Still BING++ performs consistently close to the best performance among the competitors and finally achieves 77.5\% MABO, only 3.9\% smaller than selective search. As shown in \cite{zitnickedge}, considering overall achievement this small difference is negligible in terms of proposal quality. 

We also conduct the object detection task to measure the impact of DR, ABO and MABO of proposals on real applications, and list the results in Table \ref{tab:det_full}. We run different algorithms to generate proposals and feed them to fast R-CNN \cite{girshick2015fast} with pre-trained VGG-16 model \cite{simonyan2014very} to perform detection. In terms of mean average precision (mAP), the overall detection performance of each competitive proposal algorithm is quite close to each other, \ie selective search, GoP, EdgeBoxes, LPO, MTSE-BING, BING++ and its siblings. Also the average precision (AP) for each class from BING++ is close to the best performance among the competitors. We emphasize that the running time of BING++ is about 3ms per image only using CPUs on our server, which is much faster than any of the existing competitive proposal algorithms with good quality.

To better view the difference between different proposal algorithms for object detection, we illustrate some results in Fig.~\ref{fig:samples}. Compared with BING, our BING++ produces more reasonable detections. Interestingly, the focuses of all the comparative algorithms are very similar even in such complex images, while the predicted bounding boxes vary. 

%

\subsection{Benchmark Comparison (II): Microsoft COCO}

\begin{figure*}[t]
\begin{minipage}[b]{0.245\linewidth}
 \begin{center}
 \centerline{\includegraphics[width=1.05\columnwidth]{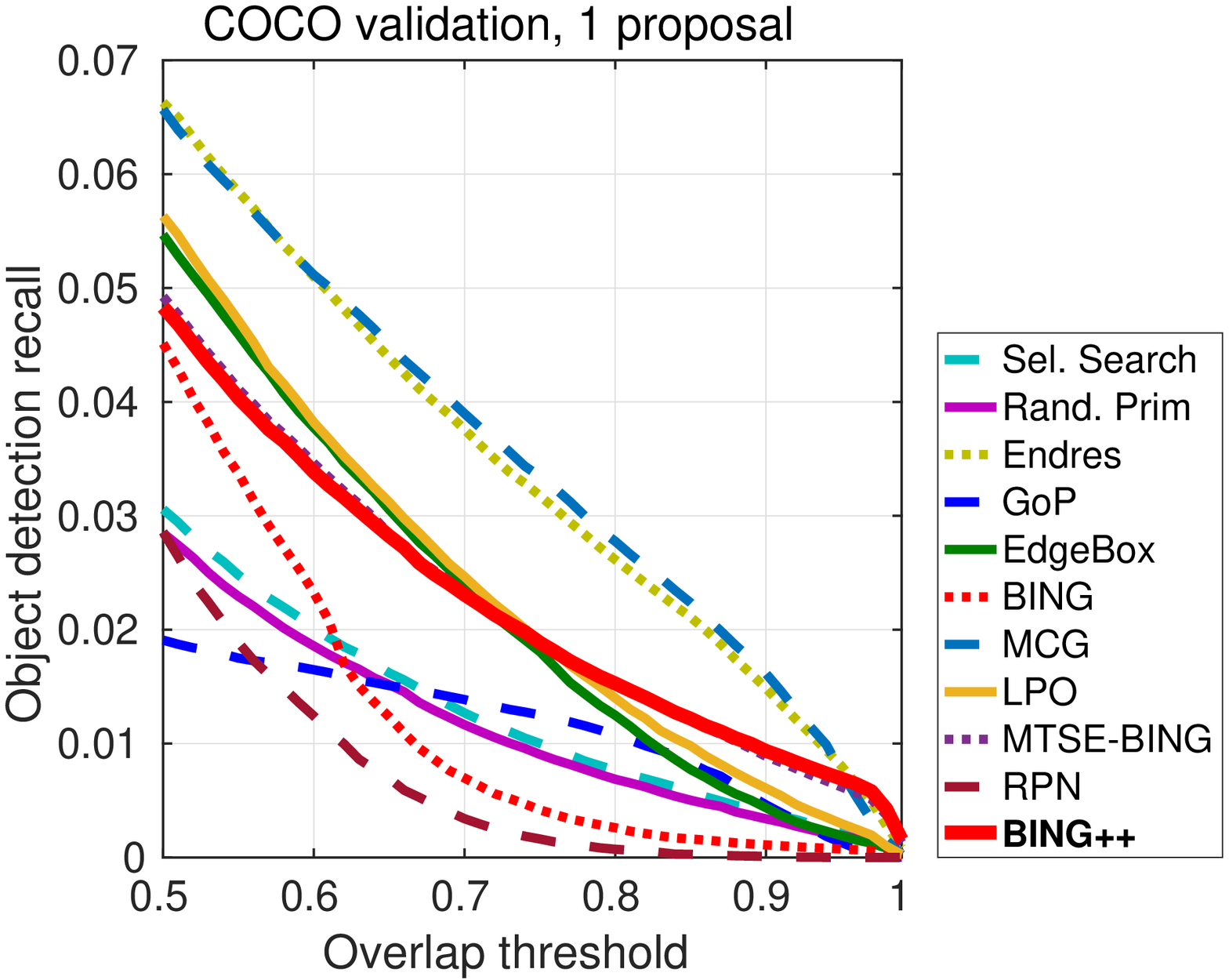}}
 \end{center}
\end{minipage}
\begin{minipage}[b]{0.245\linewidth}
 \begin{center}
 \centerline{\includegraphics[width=1.05\columnwidth]{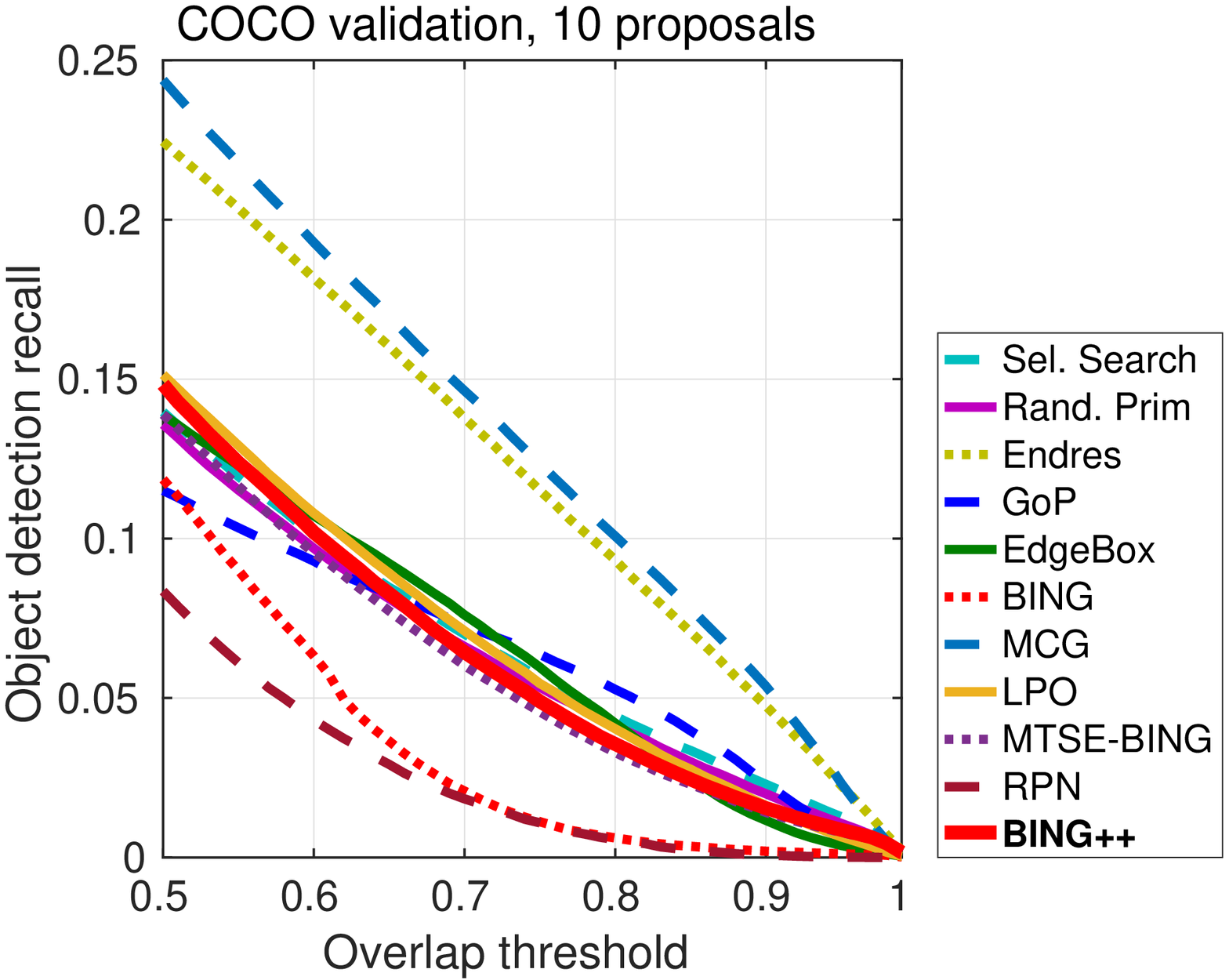}}
 \end{center}
\end{minipage}
\begin{minipage}[b]{0.245\linewidth}
 \begin{center}
 \centerline{\includegraphics[width=1.05\columnwidth]{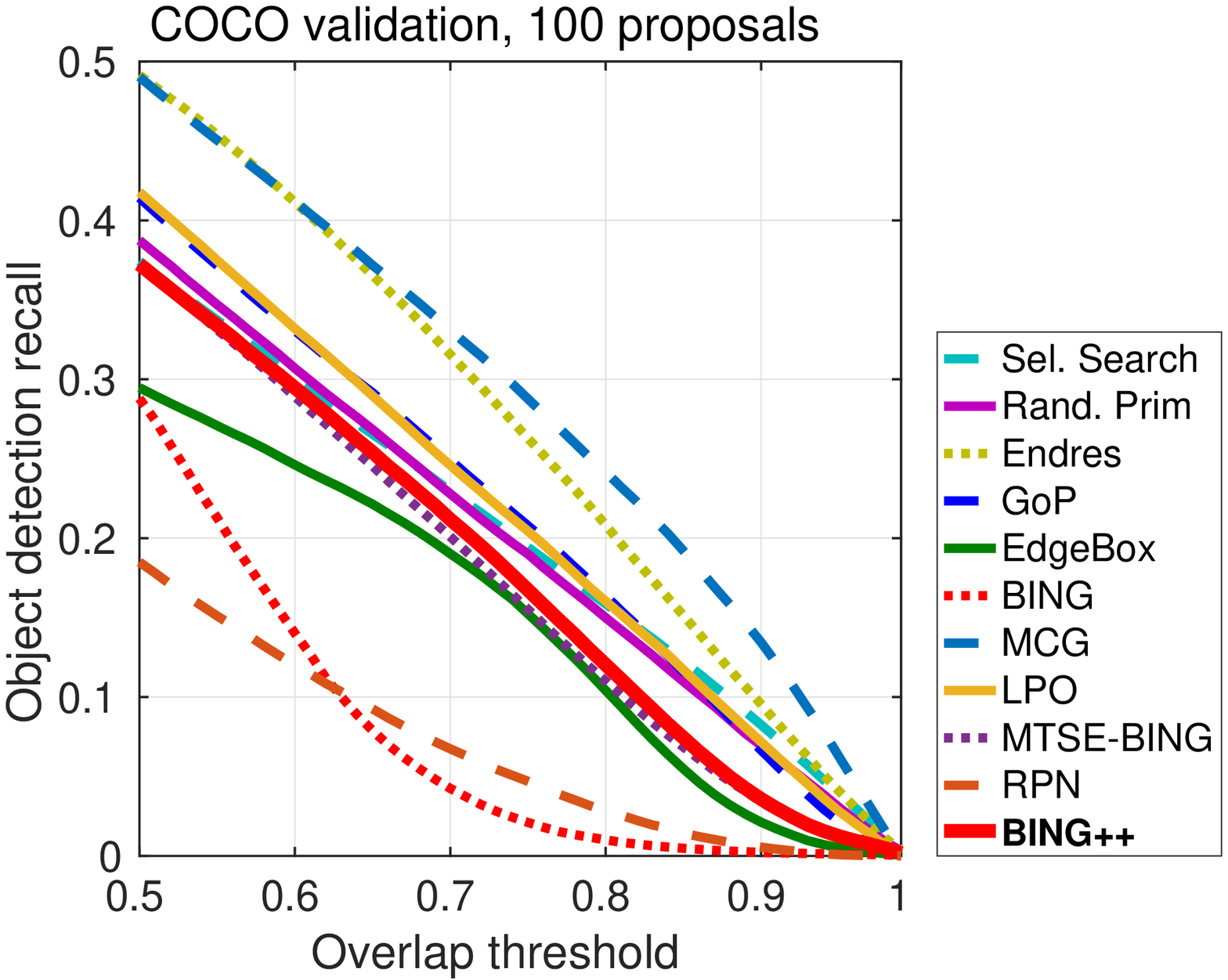}}
 \end{center}
\end{minipage}
\begin{minipage}[b]{0.245\linewidth}
 \begin{center}
 \centerline{\includegraphics[width=1.05\columnwidth]{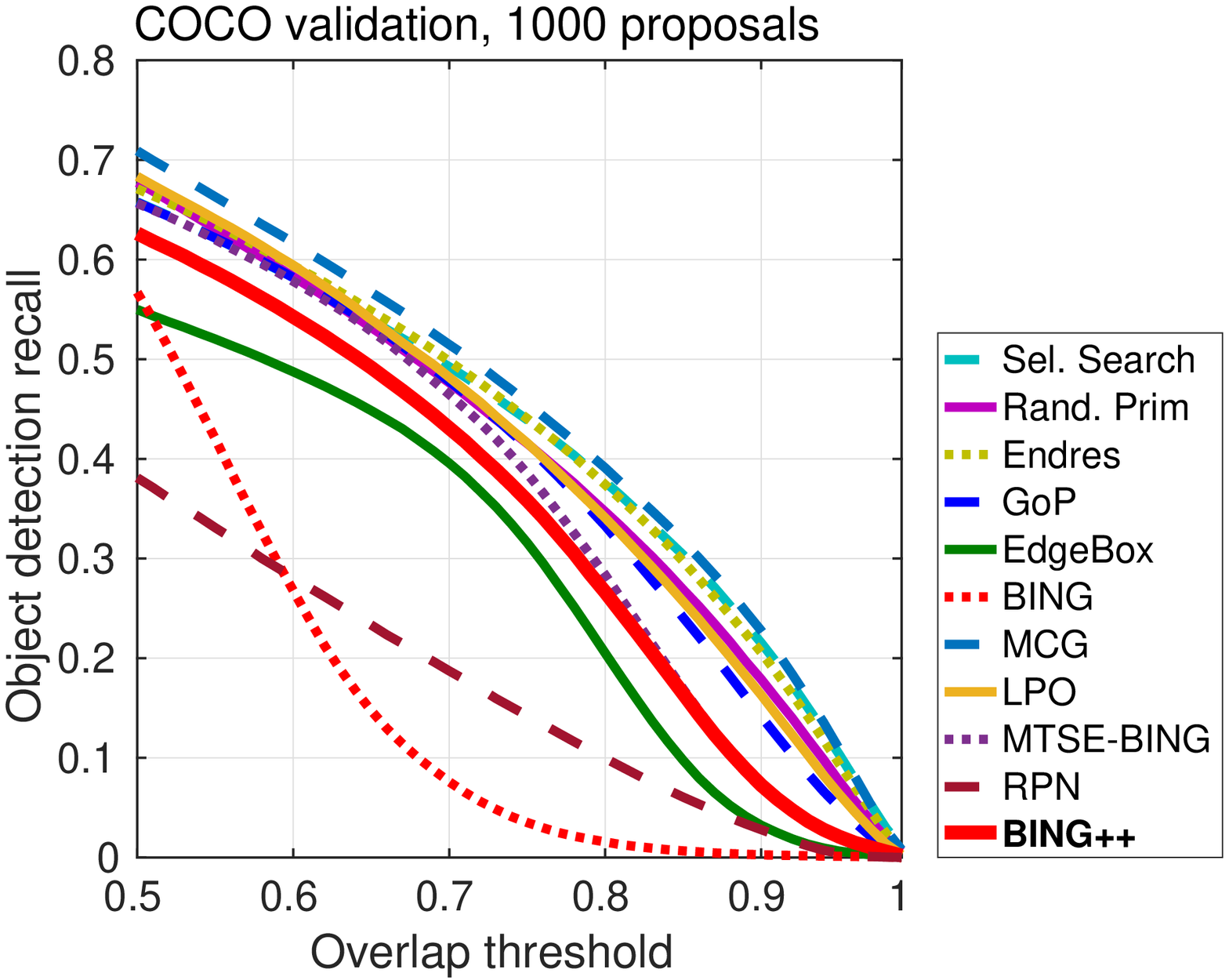}}
 \end{center}
\end{minipage}
\vspace{-9mm}
\caption{\footnotesize{Comparison of recall-overlap curves using different methods and numbers of proposals on MS COCO validation set.}}\label{fig:coco-recall-overlap}
\vspace{-5mm}
\end{figure*}

Recall that on COCO for all the methods we keep using the {\em same} (predefined or learned) parameters as those on VOC2007 without any retraining. Because there are at least 60 different object classes between the two datasets, in such way we can compare the generalization ability of different methods for the purpose of generic object proposals across different object classes/datasets.

As on VOC2007, we first compare different algorithms using DR \vs IoU overlap threshold in Fig. \ref{fig:coco-recall-overlap}. Due to the large size of the dataset, here we only compare several relatively efficient algorithms. Again BING++ performs reasonably well among the top for all the cases. Interestingly EdgeBoxes seems to struggle on COCO. One possible reason is that images in COCO are more complex than those in VOC2007, in general, leading to noisy/missing edges which confuse EdgeBoxes. Another possible reason is that EdgeBoxes is quite sensitive to its parameters, and we need to re-tune its parameters using training data in COCO. Similarly the pre-trained RPN using VOC2007 data does not work well on COCO, indicating that the RPN method is not suitable for the purpose of generic object proposal generation as it is sensitive to the parameters. However, our BING++ is more robust to parameter settings. We list in Table~\ref{tab:coco} the corresponding numbers in Fig.~\ref{fig:coco-recall-overlap} as well as MABO for numerical comparison. Note that compared with BING, both DR and MABO of BING++ are boosted significantly. 
Also in Fig.~\ref{fig:dr}(b) we show the behavior of different algorithms using DR \vs number of proposals where BING++ performs slightly worse. We speculate that this is because a large portion of small objects occurring in COCO worsen the performance, as we show in Fig. \ref{fig:dist}. Using 1,000 proposals the BING++ performs inferiorly to the best (\ie MCG), but with about 1,000 times faster. 
\begin{table}[t]\centering\footnotesize
\caption{\footnotesize{DR and MABO comparison (\%) on COCO validation dataset using the same (learned) parameters on VOC2007 for each method.}} \label{tab:coco} \vspace{-3mm}
\setlength\tabcolsep{1.8pt}
\begin{tabular}{|c|cccc|cccc|c|}
\hline
\multirow{2}{*}{Methods} & \multicolumn{4}{c|}{\# Prop., $\eta=0.5$} & \multicolumn{4}{c|}{\# Prop., $\eta=0.7$} & MABO\\
& 1 & 10 & 100 & 1000 & 1 & 10 & 100 & 1000 & (1000)\\
\hline\hline
Sel. Search \cite{Uijlings13} & 3.1 & 14.0 & 37.4 & 65.8 & 1.3 & 7.0 & 23.0 & 49.2 & 60.8 \\
Rand. Prim \cite{Manen2013iccv} & 2.9 & 13.6 & 38.7 & 67.7 & 1.2 & 6.6 & 22.8 & 47.7 & 61.8 \\
Endres \cite{Endres2012} & \underline{\bf 6.6} & 22.5 & \underline{\bf 49.2} & 67.1 & 3.8 & 13.8 & 31.6 & 49.8 & 61.0 \\
GoP \cite{krahenbuhl2014geodesic} & 1.9 & 11.5 & 41.4 & 65.7 & 1.4 & 7.3 & 24.9 & 47.8 & 59.2 \\
EdgeBox \cite{zitnickedge} & 5.5 & 13.9 & 29.5 & 55.0 & 2.4 & 7.6 & 19.0 & 39.6 & 50.0 \\        
BING \cite{BingObj2014} & 4.5 & 11.9 & 28.8 & 56.7 & 0.7 & 2.1 & 4.3 & 7.6 & 48.0 \\ 
MCG \cite{Arbelaez_CVPR14} & \underline{\bf 6.6} & \underline{\bf 24.4} & 49.0 & \underline{\bf 70.9} & \underline{\bf 3.9} & \underline{\bf 14.6} & \underline{\bf 33.1} & \underline{\bf 51.5} & \underline{\bf 64.8} \\
LPO \cite{kk-lpo-15} & 5.6 & 15.1 & 41.8 & 68.4 & 2.5 & 7.1 & 24.6 & 48.2 & 61.3 \\
MTSE-BING \cite{wang2015improving} & 4.9 & 13.9 & 37.3 & 65.7 & 2.3 & 6.0 & 20.1 & 46.5 & 58.0 \\
RPN \cite{ren2015faster} & 2.9 & 8.4 & 18.5 & 38.1 & 0.3 & 1.8 & 6.7 & 18.7 & 25.4 \\
\hline\hline
{\bf BING++} & 4.8 & 14.8 & 37.2 & 62.6 & 2.3 & 6.4 & 21.2 & 43.1 & 56.0 \\
\hline
\end{tabular}
\end{table}

\begin{figure}[t]
\begin{minipage}[b]{0.495\columnwidth}
 \begin{center}
 \centerline{\includegraphics[width=1.05\columnwidth]{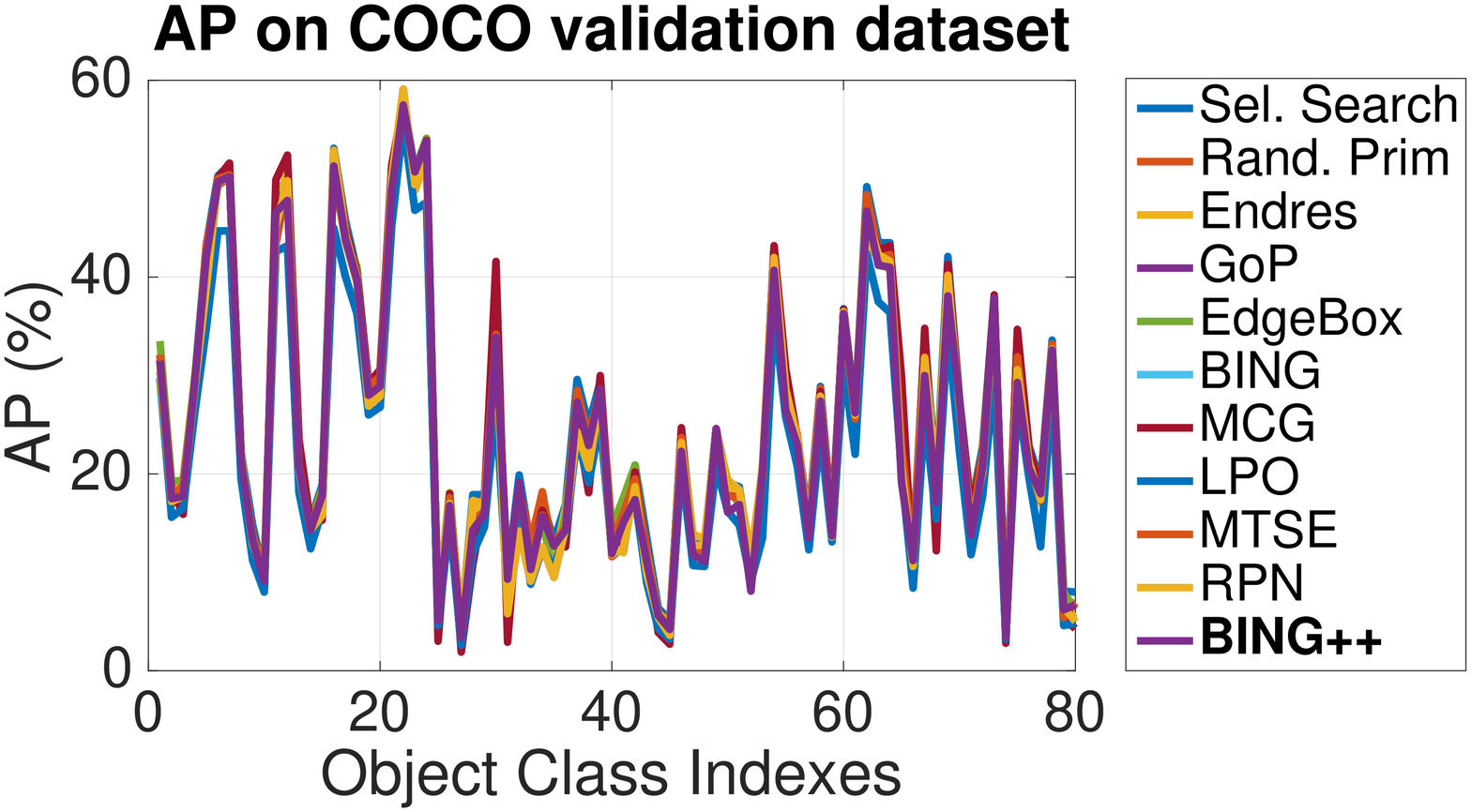}}
 \centerline{\footnotesize{(a)}}
 \end{center}
\end{minipage}
\begin{minipage}[b]{0.495\columnwidth}
 \begin{center}
 \centerline{\includegraphics[width=1.05\columnwidth]{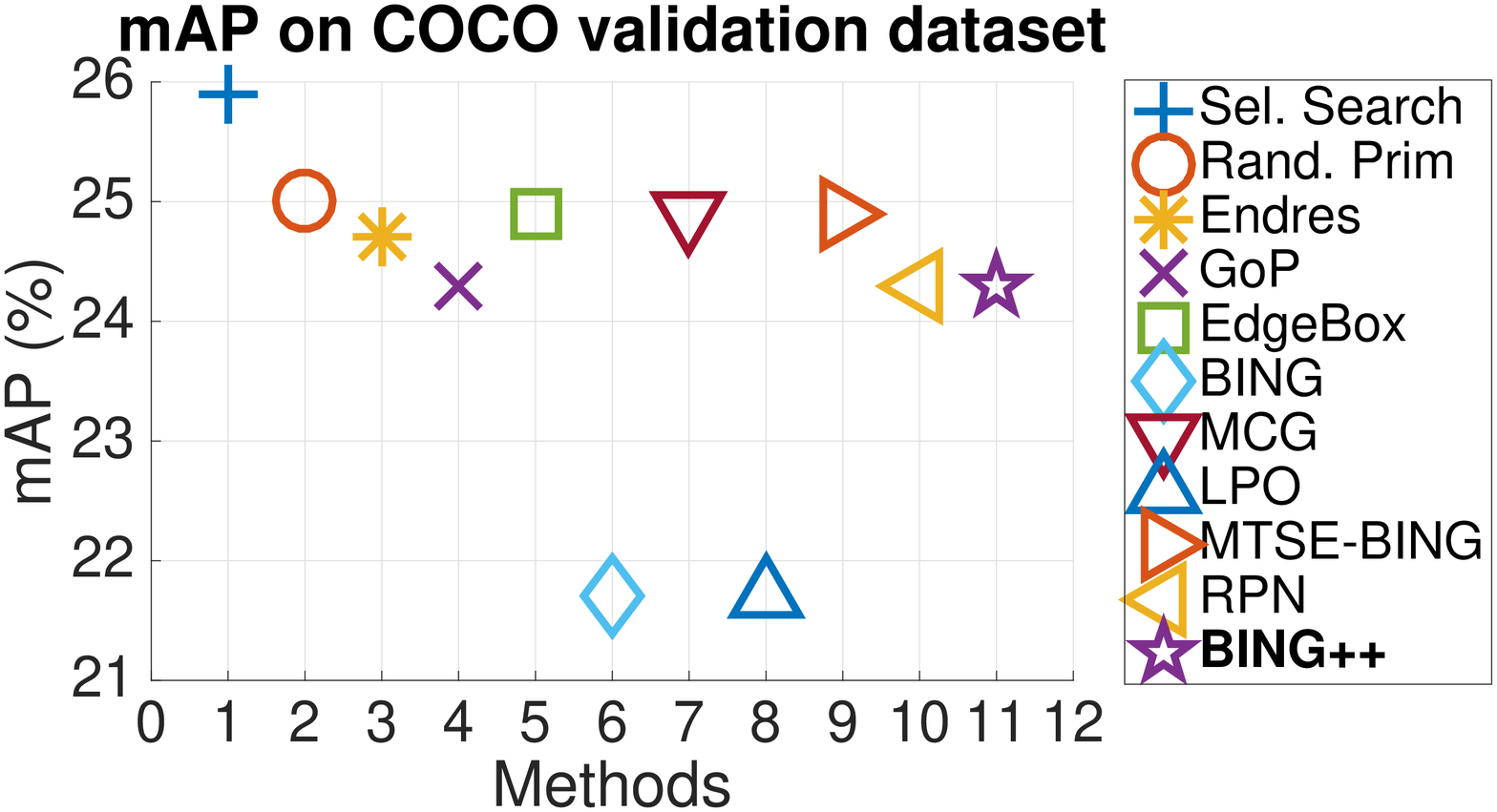}}
 \centerline{\footnotesize{(b)}} 
 \end{center}
\end{minipage}
\vspace{-9mm}
\caption{\footnotesize{{\bf (a)} AP and {\bf (b)} mAP comparison for object detection on COCO validation dataset.}}\label{fig:det_coco}
\vspace{-3mm}
\end{figure}

We also show the AP and mAP scores for object detection on COCO validation dataset in Fig. \ref{fig:det_coco} using fast R-CNN with pre-trained VGG-16 model. Different from VOC2007, all the methods on AP in Fig. \ref{fig:det_coco}(a) behave similarly with marginal gaps. In terms of mAP shown in Fig. \ref{fig:det_coco}(b) BING++ performs comparably with the best method (\ie selective search) with difference of 1.6\%. Interestingly MCG, the best method in Table \ref{tab:coco}, performs slightly worse than selective search, but better than BING++ with margin of 0.6\%. LPO is better than BING++ in terms of MABO in Table \ref{tab:coco}, but worse in terms of mAP. These observations suggest that there is no clear positive correlation between proposal quality and detection performance, but in general better proposal quality will probably lead to better detection performance.

%

\section{Conclusion}
In this paper, we propose a novel object proposal algorithm, BING++, for generating high quality proposals efficiently based on BING. BING is a simple and fast object proposal algorithm with only a few atomic (\eg {\sc add}, {\sc bitwise}, \etc) operations for scoring each window in images. However, as we examined in Fig.~\ref{fig:abo-stats} and Fig.~\ref{fig:stats}, the proposal localization quality generated by BING is not satisfactory, on average.

Our BING++ algorithm essentially improves the proposal quality of BING significantly as well as preserves its high computational efficiency. BING++ consists of three components sequentially: (1) BING algorithm, (2) a novel EdgeRecursiveBox algorithm, (3) another novel SegmentRecursiveBox algorithm, where (2) and (3) manage to adapt the BING proposals to object boundaries. We leverage the facts that edges/segments in images can be used to approximate object boundaries, and the ground-truth bounding boxes should cover the entire objects as tightly as possible. Based on these considerations, we propose estimating ground-truth bounding boxes in a recursive way using nearest edge points first and then segments, based on which we update the proposals to the ones that cover the sets of pixels of nearest edges or segments tightly. 

Fundamentally our RecursiveBox algorithm essentially tries to solve a sequential minimization problem with 0/1-loss function for object proposal generation. Due to high non-convexity of our problem, we propose a novel parameter space quantization algorithm as a solver to exhaustively search for an approximate solution. Comprehensive experiments on VOC2007 and Microsoft COCO have demonstrated the effectiveness and efficiency of our algorithm, making BING++ among the top object proposal generation algorithms with achievement of better trade-off between proposal quality and computational efficiency in the literature.
%

As BING++ predicts a small set of object bounding boxes based on edge and segmentation information, it may suffer similar limitations as other edge or segmentation based proposal algorithms (see proposal repeatability section in \cite{Hosang2015Pami}). However, empirically these two types of information are actually working complementarily against the drawbacks of each other, resulting in better proposals in terms of localization/detection quality as we demonstrated in our comprehensive comparison.

The major limitation of BING++ is that it works inferiorly with small objects sometimes, as we show in Fig. \ref{fig:dist}. This is probably because BING cannot generate reasonably good proposals to cover these small objects, or no edges or segments are produced for them, leading to the failure cases. Recently this problem of small object proposal generation has attracted the attention of researchers \cite{jie2016scale}, and it will be one of our future works to further improve proposal quality as well as preserving computational efficiency. 

Our BING++ could be one possible solution (partially) to the computational bottleneck in real-time object detection with thousands of object categories by efficiently reducing the number of bounding boxes as proposals that are needed to be verified. Another effort 
could be reducing the classification complexity such as \cite{dean2013fast}. It would be very interesting to explore the integration of BING++ with such methods for detection in the future. Because of the high efficiency of BING++, it is also desirable in many video applications such as person re-identification \cite{zhang_eccv14,zhang_iccv15_reid}, video event detection \cite{Castanon_mm15}, and weakly supervised learning \cite{Li2016}. We will explore such applications in the future as well.

\section*{Acknowledgement}
We thank Tolga Bolukbasi from Boston University for helping to prepare some of experimental results.

\ifCLASSOPTIONcaptionsoff
  \newpage
\fi



\bibliographystyle{IEEEtran}\footnotesize
\bibliography{egbib}
\vspace{-15mm}
\begin{biography}[{\includegraphics[width=1in,height=1.25in,clip,keepaspectratio]{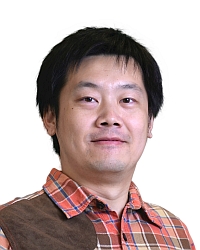}}]{Ziming Zhang}
is a research scientist at Mitsubishi Electric Research Laboratories (MERL). Before joining MERL he was a research assistant professor at Boston University. He received his PhD degree in 2013 from Oxford Brookes University, UK, under the supervision of Prof. Philip H. S. Torr. His research areas include object recognition and detection, machine learning, optimization, large-scale information retrieval, visual surveillance, and medical imaging analysis. His works have appeared in TPAMI, CVPR, ICCV, ECCV, ACM Multimedia and NIPS.
\end{biography}
\vspace{-15mm}
\begin{biography}[{\includegraphics[width=1in,keepaspectratio]{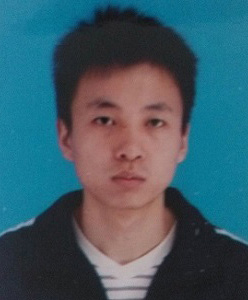}}]{Yun Liu}
is a Ph.D. Candidate with College of Computer Science and Control Engineering, Nankai University. He received his bachelor degree from Nankai University in 2016. His research interest includes computer vision and machine learning (especially deep learning).
\end{biography}
\vspace{-11mm}
\begin{biography}[{\includegraphics[width=1in,height=1.25in,clip,keepaspectratio]{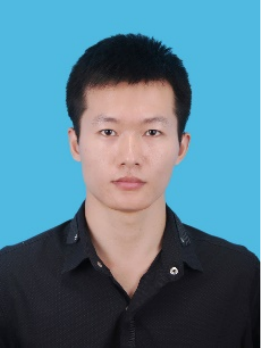}}]{Xi Chen}
received his B.S. degree in the Automation School, Huazhong University of Science and Technology (HUST) in 2015. He is currently a M.S. in the Automation School, HUST. He won a First Prize in the national finals of ``Freescale'' (now called ``NXP'') Cup National University Students Intelligent Car Race in 2014. His research interests include object detection, object tracking, human-computer interaction and machine learning.
\end{biography}
\vspace{-11mm}
\begin{biography}[{\includegraphics[width=1in,height=1.25in,clip,keepaspectratio]{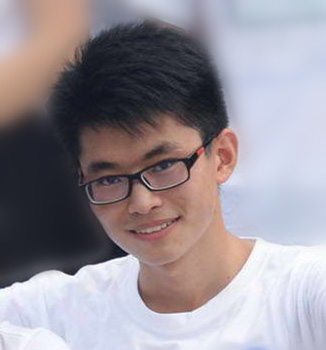}}]{Yanjun Zhu}
received his B.S. degree from Huazhong University of Science and Technology (HUST), China, in 2014. He is currently working toward the M.S. degree in the School of Automation, HUST. His research interests include visual object detection, weakly supervised learning and the application of computer vision in agriculture.
\end{biography}
\vspace{-11mm}
\begin{biography}[{\includegraphics[width=1in,keepaspectratio]{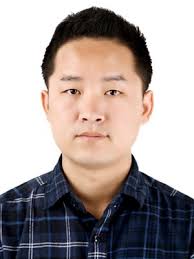}}]{Ming-Ming Cheng}
received his PhD degree from Tsinghua University in 2012. Then he did 2 years research fellow, with Prof. Philip Torr in Oxford. He is now an associate professor at Nankai University, leading the Media Computing Lab. His research interests includes computer graphics, computer vision, and image processing. He is on the program committees of several international conferences on computer vision and computer graphics, and reviews papers regularly for journals and conferences including the IEEE TPAMI, ACM TOG, ACM SIGGRAPH, IEEE CVPR, etc.
\end{biography}
\vspace{-11mm}
\begin{biography}[{\includegraphics[width=1in,height=1.25in,clip,keepaspectratio]{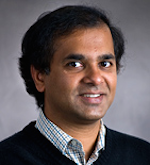}}]{Venkatesh Saligrama}
is a faculty member in the Department of Electrical and Computer Engineering and Department of Computer Science (by courtesy) at Boston University. He holds a PhD from MIT. His research interests are in Statistical Signal Processing, Machine Learning, Vision \& Learning, Information and Decision theory. He has edited a book on Networked Sensing, Information and Control. He has served as an Associate Editor for IEEE Transactions on Information Theory, IEEE Transactions on Signal Processing and has been on Technical Program Committees of several IEEE conferences. He is the recipient of numerous awards including the Presidential Early Career Award (PECASE), ONR Young Investigator Award, and the NSF Career Award. More information about his work is available at \url{http://sites.bu.edu/data}.
\end{biography}
\vspace{-11mm}
\begin{biography}[{\includegraphics[width=1in,height=1.25in,clip,keepaspectratio]{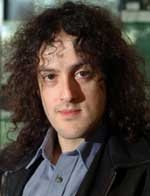}}]{Philip H.S. Torr}
received the PhD degree from Oxford University. After working for another three years at Oxford, he worked for six years as a research scientist for Microsoft Research, first in Redmond, then in Cambridge, founding the vision side of the Machine Learning and Perception Group. He is now a professor at Oxford University. He has won awards from several top vision conferences, including ICCV, CVPR, ECCV, NIPS and BMVC. He is a Royal Society Wolfson Research Merit Award holder.
\end{biography}

\vfill

\end{document}